\documentclass[3p,12pt]{elsarticle}

\usepackage{amssymb}

\usepackage{hhline}
\usepackage{threeparttable}
\usepackage{multirow}
\usepackage{xcolor}
\usepackage{booktabs}
\usepackage{multicol}
\usepackage[ruled,vlined]{algorithm2e}
\usepackage[figuresright]{rotating} 
\usepackage{subfigure}
\usepackage{graphicx}
\usepackage{colortbl}
\usepackage{color}

\begin{document}
	
	\begin{frontmatter}
		
		\title{\textcolor[rgb]{0,0,0}{RL-GA: A Reinforcement Learning-Based Genetic Algorithm for Electromagnetic Detection Satellite Scheduling Problem} }
		
		\author[]{Yanjie Song\corref{cor1}\fnref{fn1}}
		\ead{songyj\_2017@163.com}
		\author[]{Luona Wei\fnref{fn1}}
		\ead{wlnelysion@163.com}
		\author[]{Qing Yang\fnref{fn1}}
		\ead{yangqing@163.com}
		\author[]{Jian Wu\fnref{fn1}}
		\ead{1551699723@qq.com}
		\author[]{Lining Xing\fnref{fn2}}
		\ead{xinglining@gmail.com}
		\author[]{Yingwu Chen\fnref{fn1}}
		\ead{ywchen@nudt.edu.cn}

		\cortext[cor1]{Corresponding author}
		\fntext[fn1]{College of Systems Engineering, National University of Defense Technology, Changsha, Hunan, China, 410073}
		\fntext[fn2]{School of Electronic Engineering, Xidian University, Xi’an, China, 710126}

		\begin{abstract}
			\textcolor[rgb]{0,0,0}{The study of electromagnetic detection satellite scheduling problem (EDSSP) has attracted attention due to the detection requirements for a large number of targets. This paper proposes a mixed-integer programming model for the EDSSP problem and a genetic algorithm based on reinforcement learning (RL-GA). Numerous factors that affect electromagnetic detection are considered in the model, such as detection mode, bandwidth, and other factors. The RL-GA embeds a Q-learning method into an improved genetic algorithm, and the evolution of each individual depends on the decision of the agent. Q-learning is used to guide the population search process by choosing evolution operators. In this way, the search information can be effectively used by the reinforcement learning method. In the algorithm, we design a reward function to update the Q value. According to the problem characteristics, a new combination of $<$state, action$>$ is proposed. The RL-GA also uses an elite individual retention strategy to improve search performance. After that, a task time window selection algorithm (TTWSA) is proposed to evaluate the performance of population evolution. Several experiments are used to examine the scheduling effect of the proposed algorithm. Through the experimental verification of multiple instances, it can be seen that the RL-GA can solve the EDSSP problem effectively. Compared with the state-of-the-art algorithms, the RL-GA performs better in several aspects.}
			
		\end{abstract}
		
		\begin{keyword}
			\textcolor[rgb]{0,0,0}{reinforcement learning, electromagnetic detection, scheduling, genetic algorithm, learning adaptive, Q-Learning}
		\end{keyword}
	\end{frontmatter}
	
	\section{Introduction}
	In recent years, the rapid development of aerospace has provided new solutions for various types of tasks such as information communication, environmental monitoring, and disaster forecasting\cite{b1}. The applications that satellites can play can be divided into three categories: observation, communication, and navigation according to the different carrying payloads. \textcolor[rgb]{0,0,0}{Here, observation satellites can be further classified according to the payload equipped with visible light, infrared, synthetic aperture radar, antenna, and others.} Satellites that use signal receivers and antennas as payloads are called electromagnetic detection satellites. Electromagnetic detection satellites can detect and process electromagnetic signals to obtain helpful information. Compared with optical imaging satellites, electromagnetic detection satellites have a wider detection range and are not easily affected by weather factors. \textcolor[rgb]{0,0,0}{As a result, it can execute tasks in all types of weather conditions. Based on known signal characteristics, electromagnetic detection satellites can also search for unknown signals through a wide range of frequencies.}
	
	\textcolor[rgb]{0,0,0}{There are a large number of electromagnetic signals on the surface of the earth, and many types of electromagnetic signals constitute the surface electromagnetic environment together.} The detection of surface electromagnetic environments can effectively support work in various fields such as environment, agriculture, military, meteorology, and others. The electromagnetic detection satellite scheduling problem (EDSSP) is to plan the satellite resources used to execute the demand and determine the specific execution time of each task. \textcolor[rgb]{0,0,0}{Detection demands are put forward by various users and are expected to be met fully. However, the tasks that can be performed during each orbit are quite limited. This is due to the existence of limitations such as satellite orbit, detection range, and operating conditions.} The goal of electromagnetic detection satellite scheduling is to generate a reasonable task sequence within a certain time range. \textcolor[rgb]{0,0,0}{This sequence is expected to generate sufficient detection profit and satisfy users' preferences. While the characteristic of oversubscription is very common in a satellite scheduling problem. How to build a model and design a scheduling algorithm then becomes the most central issue.}
	
	\textcolor[rgb]{0,0,0}{Electromagnetic signal detection needs to effectively balance the relationship between the detection accuracy and the satellite capability. From the signal detection activity itself, when the density of detection frequency points is increased to improve the detection accuracy, the amount of data generated will also increase accordingly. The capacity of the onboard storage device is limited. If the storage device is fully used, it will not be possible to execute a new task until it is erased. The reasonable setting of detection accuracy according to user needs can improve detection performance.}
	
	\textcolor[rgb]{0,0,0}{Existing studies of electromagnetic detection satellite mission planning have proposed corresponding solution methods for multiple scenarios such as multiple area detection and moving target detection\cite{b2,b3}. However, the specific modes of satellites are treated in a simplified way in these studies, which only addresses the problem that is common to all types of satellite scheduling problems. It can be said that these studies are not mature and in-depth enough. Compared with the electromagnetic detection satellite scheduling problem, there are many studies related to the optical satellite observation scheduling problem\cite{b4,b5}.} In terms of model building, Berger and Barkaoui used the quadratic programming method to construct the visible light satellite scheduling model\cite{b6,b7}. Cho et al. proposed a two-step binary linear programming model to formulate a satellite observation scheme for a low-orbit satellite constellation\cite{b8}. Niu et al. considered using satellites to complete post-disaster large-area observation tasks built a multi-objective optimization model and used genetic algorithms to solve this problem\cite{b9}. Chen et al. proposed a mixed-integer programming model for the agile satellite scheduling problem, constructing decision variables based on conflicts between time windows and using 5-index variables to obviate the need for a Big-M approach\cite{b10}.
	
	Evolutionary algorithms have been widely used in solving satellite task scheduling problems. The applications of the genetic algorithm\cite{b11}, differential evolution\cite{b12}, ant colony algorithm\cite{b13}, memetic algorithm\cite{b14} and others algorithms have increased rapidly in recent years. Compared with accurate algorithms, the evolutionary algorithm has a lower computational cost on large-scale problems. The studies of \cite{b15,b16} show that genetic algorithms are more suitable for solving satellite task scheduling problems than other evolutionary algorithms. E. et al. proposed an individual reconfiguration-based integer coding genetic algorithm (IRICGA) for regional target observation planning and designed a new area partitioning method based on two kinds of discrete parameters\cite{b17}. Zhang and Xing adopted a novel idea of encoding and decoding for the integrated satellite imaging and data transmission scheduling problem to improve the search efficiency of the genetic algorithm. Two test cases are used to verify the algorithm effect \cite{b18}. Song et al. and Du et al. used non-dominated sorting-based and decomposition-based multi-objective evolutionary algorithms, respectively, to find high-quality solutions for satellite range scheduling problems\cite{b19,b20}. Xhafa et al. proposed a genetic algorithm combined with the STK solution tool to obtain a plan of ground stations\cite{b21}.
	
	Existing research on reinforcement learning to solve satellite scheduling problems rarely considers combining it with other algorithms. \textcolor[rgb]{0,0,0}{Solving this problem using the reinforcement learning method alone is also just beginning. Wei et al. directly used reinforcement learning methods to solve imaging satellite scheduling problems\cite{b22}.} Chen et al. built an end-to-end reinforcement learning framework, used the attention mechanism, and proposed an actor-critic network training method\cite{b23}. \textcolor[rgb]{0,0,0}{To the best of our knowledge, there is no research that the embedded approach to solving the satellite scheduling problem.}
	
	Combining reinforcement learning with search algorithms can effectively utilize the respective advantages of the two methods to improve search efficiency and search performance \cite{b24}. A part of the related research focuses on numerical optimization problems\cite{b25,b26}. 
	
	\textcolor[rgb]{0,0,0}{Rodríguez-Esparza et al. \cite{b27} used a hyper-heuristic algorithm consisting of a multi-armed bandit RL method and a simulated annealing algorithm to solve the path planning problem for electric vehicles with capacity constraints. This hyper-heuristic algorithm uses RL to select a neighborhood search strategy and obtains a legal route by repairing the solution. Zhang et al. \cite{b28} proposed an improved algorithm combining the Q-learning method and multi-objective particle swarm algorithm for the multi-UAV path planning problem. This algorithm uses a deep reinforcement learning method to determine the mode used for particle swarm search. Du et al \cite{b29} considered the effect of electricity cost in a flexible job shop scheduling problem and proposed a reinforcement learning-based distribution estimation algorithm. The algorithm uses DQN to select the rules for process adjustment. Half of the individuals are constituted in this way and the other half is optimized by the distribution estimation algorithm.}
	
	\textcolor[rgb]{0,0,0}{These combined methods above achieve good performance and demonstrate the advantages of this algorithm design idea. For the EDSSP problem, we try to use a new combination of a genetic algorithm and reinforcement learning method to find the ideal plan. Control parameters are extremely important for each evolutionary algorithm, and these parameters are always sensitive to the problem. This easily affects the generalization of the algorithm. The reinforcement learning method has strong generalization capabilities but is prone to poor performance due to a lack of domain-specific knowledge. So, we try to apply the reinforcement learning method to the evolutionary algorithm and effectively combine the respective advantages of these two methods.}
	
	\textcolor[rgb]{0,0,0}{Our proposed reinforcement learning-based genetic algorithm allows researchers to invest more effort in algorithm and strategy design. The reinforcement learning approach helps the genetic algorithm find higher-quality solutions through intelligent decision-making. Such an approach has strong application prospects and can effectively deal with various types of complex situations. For the method, this is the first time that reinforcement learning is used to embed into an evolutionary algorithm for solving a satellite scheduling problem, and this research idea can also be applied to other combinatorial optimization problems.}
	
	The main contributions of our research are:
	
	1. A refined model of the EDSSP problem is built. \textcolor[rgb]{0,0,0}{This mixed-integer programming model considers many practical factors such as detection mode, detection angle, and data volume, which is conducive to obtaining a more practical scheduling scheme.}
	
	\textcolor[rgb]{0,0,0}{2. A genetic algorithm based on reinforcement learning is proposed. The algorithm uses a reinforcement learning method to drive the population search. Combining the problem characteristics of EDSSP, we construct a new combination of state and action methods and propose a reward function and strategy selection method. In addition, a two-stage task time window selection algorithm is proposed. It is used to generate detection schemes and calculate the fitness function values.}
	
	\textcolor[rgb]{0,0,0}{3. The effectiveness of the proposed RL-GA is verified by extensive experiments. The RL-GA is excellent in terms of solution performance and can solve large-scale  EDSSP effectively. It also provides an idea for solving other combinatorial optimization problems.}
	
	\textcolor[rgb]{0,0,0}{The remainder of this paper is organized as follows. The second part introduces the model of the electromagnetic detection satellite scheduling problem. The third part introduces the reinforcement learning-based genetic algorithm and the task time window selection algorithm. The fourth part verifies the effect of the proposed algorithm. The fifth part summarizes the content and analyzes possible directions for further research in the future.}
	\section{Model}
	\subsection{Problem Description}
	The EDSSP problem is to designate a time-ordered task execution sequence for electromagnetic detection satellites. \textcolor[rgb]{0,0,0}{The goal is to maximize the detection sequence profit while satisfying various satellite constraints. For a satellite to successfully perform any mission, it needs to determine the on/off time of the receiver i.e. the start time and the end time of the mission. A series of parameter settings such as detection mode, frequency, bandwidth, and polarization mode must also be followed.}
	
	The time range from the beginning to the end of the signal beam coverage of the electromagnetic satellite is called the visible time window. Since the electromagnetic satellite antenna can effectively detect a wide range of ground signals, to reduce noise and improve the detection accuracy, the angle between the signal source and the pointing direction of the satellite antenna needs to be within a certain range.
	
	\textcolor[rgb]{0,0,0}{The detection quality is affected by two factors. On the one hand, the signal gain is affected by the angular relationship between the satellite antenna and the signal source. When the center of the satellite antenna passes directly above the signal source, the maximum signal gain can be obtained, and the signal gain is directly linked to the detection profit. Signal gain is inversely related to the angle between the antenna and the signal source. In other words, when the centerline of the signal source beam coincides with the extension line of the satellite antenna pointing direction, the detection effect is the best. On the other hand, the detection accuracy will also affect the detection profit. The detection accuracy is limited by the inherent capability of the receiver, and the number of frequency points will also have a direct impact on the detection accuracy. There is a positive correlation between the bandwidth and the number of frequency points, and the amount of data obtained by detection varies significantly depending on settings. The limited satellite storage capacity means that only a fraction of tasks can be set to the highest bandwidth, while many tasks need to be set to a smaller bandwidth.}
	
	\subsection{Symbols and Variables}
	$T$ : Set of tasks, a total of ${\left| T \right|}$ tasks. For task $task_j$, the following attributes are defined:
	
	$es{t_j}$: The earliest available start time of the task.
	
	$le{t_j}$: The latest available end time of the task.
	
	$du{r_j}$: Duration of the task. 
	
	$\theta _j^{\max }$: Maximum allowable detection angle of the task.
	
	$degre{e_j}$ : Task importance level.
	
	${m_j}$: The amount of data for the task.
	
	${G_j}$: The signal gain can be obtained from the task.
	
	$S$: Set of satellites, a total of ${\left| S \right|}$ satellites. For satellite ${s_i}$, the following attributes are defined:
	
	$D$: Satellite antenna diameter.
	
	$\eta $ : Antenna efficiency.
	
	${O_i}$: Set of Orbits, a total of $\left| {{O_i}} \right|$ orbits belonging to satellite ${s_i}$.
	
	$\beta $ : Satellite detection unit data volume.
	
	$M$: Satellite storage capacity.
	
	${\Gamma _{pol}}$ : Satellite polarization transition time.
	
	${\Gamma _{mode}}$ : Satellite detection mode transition time.
	
	${\Gamma _{band}}$: Satellite bandwidth setting transition time.
	
	${\Gamma _{fre}}$: Satellite frequency band transition time.
	
	$\Delta $ : Satellite load on/off time.
	
	%
	%
	%
	
	$TW$ : Set of time windows, with a total of $\left| {TW} \right|$ time windows. For the time window $t{w_{ijok}}$, the following attributes are defined:
	
	$EV{T_{ijok}}$ : The earliest visible time of the task $j$ in the time window $k$ on orbit $o$ for satellite $i$.
	
	$LV{T_{ijok}}$ : The latest visible time of the task $j$ in the time window $k$ on orbit $o$ for satellite $i$.
	
	$\theta _{_{ijok}}^t$ : Detection angle  of the satellite $i$ at the time $t$ in the time window $k$ of the task $j$ on the orbit $o$.

	$I$: A big integer.
	
	Decision variables:
	
	${x_{ijok}}$ : Whether the satellite $i$ is in the time window $k$ on the orbit $o$ whether the task $j$ is executed, if the task is executed, ${x_{ijok}} = 1$; otherwise, ${x_{ijok}} = 0$.
	
	$s{t_{ijo}}$ : Start time of the satellite $i$ on the orbit $o$ to execute the task $j$.
	
	\subsection{Mathematical Model}
	\textcolor[rgb]{0,0,0}{We refer to the studies of \cite{b30} and \cite{b31} and make the following assumptions.}
	
	\textbf{Assumptions:}
	\begin{itemize}
		\item All electromagnetic detection satellites have the same receivers and storage devices;
		
		\item The detection process will not be affected by external factors;
		
		\item The detection task is definite, and there will be no temporary changes or cancellations;
		
		\item The satellite has sufficient energy during orbit;
		
		\item Each task can be completed after one detection, without repeated detection.
	\end{itemize}
	
	The calculation formula of the detection profit that can be obtained by a single detection task is:
	
	\begin{equation}
		{G_j} = {G_0} \cdot {\left[ {\frac{{{J_1}\left( u \right)}}{{2u}} + 36\frac{{{J_3}\left( u \right)}}{{{u^3}}}} \right]^2}\left( {dBi} \right)\label{eq1}
	\end{equation}
	
	\begin{equation}
		{G_0} = \eta \frac{{{\pi ^2}{D^2}}}{{{\lambda ^2}}}\left( {dBi} \right)\label{eq2}
	\end{equation}

	where $u = 2.07123\sin \left( \theta  \right)/\sin \left( {{\theta _{3dB}}} \right)$, ${{J_1}\left( u \right)}$ and ${{J_3}\left( u \right)}$ are the 1st and 3rd order Bessel functions of the first kind, respectively. $\theta$  is the angle between the satellite antenna and the center of the signal source, $\theta _{3dB}$ is the angle at which the antenna gain is attenuated by 3dB relative to the center of the beam, and the calculation formula as follows.
	\begin{equation}{\theta _{3dB}} = 70\lambda /D\end{equation}
	
	where $\lambda $ represents the wavelength, and $D$ indicates the diameter of the antenna.
	
	In this paper, the bandwidth used by satellites to perform detection tasks is dynamically matched according to the priority of detection tasks. The importance of the task is high, and the bandwidth used is large so that the detection effect will be better. However, due to the limitation of satellite storage, the detection bandwidth needs to be scientifically set. The formula for setting the bandwidth according to the degree of importance ${degre{e_j}}$ is shown below.
	\begin{equation}\varphi \left( {degre{e_j}} \right) = \left\{ {\begin{array}{*{20}{c}}
				{bandwidt{h_1}}&{degre{e_j} > 75}\\
				{bandwidt{h_2}}&{50 < degre{e_j} \le 75}\\
				{bandwidt{h_3}}&{25 < degre{e_j} \le 50}\\
				{bandwidt{h_4}}&{0 < degre{e_j} \le 25}
		\end{array}} \right.\end{equation}
	
	The bandwidth setting can also affect the detection profit, and the signal gain can be measured by the function $\Omega \left[ {\varphi \left( {degre{e_j}} \right)} \right]$. When the bandwidth of the detection task $j$ is set according to $\varphi \left( {degre{e_j}} \right)$, the amount of data generated per second is $\beta  \cdot \varphi \left( {degre{e_j}} \right)$. Combined with the task detection time $du{r_j}$, the total data amount of task $j$ can be obtained by:
	\begin{equation}{m_j} = \beta  \cdot \varphi \left( {degre{e_j}} \right) \cdot du{r_j}\end{equation}

	\textcolor[rgb]{0,0,0}{The main parameters of the electromagnetic detection task include frequency, bandwidth, polarization, and detection mode. The parameters of satellite $s_i$ for task $task_j$ are set as $\left(fre_{ij},band_{ij},pol_{ij},mode_{ij}\right)$. Where $fre_{ij}$ denotes the detection frequency, $band_{ij}$ denotes the bandwidth, $pol_{ij}$ denotes the polarization mode, and $mode_{ij}$ denotes the detection mode. The electromagnetic detection satellite needs to adjust the parameters of the onboard equipment when performing different tasks. The transition time between tasks is composed of four parts. The first part is the time required for the change of the polarization mode. The second part is the time required for the change of the detection mode. The third part is the time required for the change of the frequency, and the fourth part is the time required for the change of the bandwidth. In addition, it also takes a certain amount of time for each onboard equipment to be turned on and off, and this time interval $\delta$ must be satisfied between every two tasks. To simplify the constraint judgment, we introduce a new variable $tra{n_{ijj'}}$, which represents the total transition time. The transition time of the two tasks $j$ and ${j'}$ is as follows.}
	\begin{equation}
		\begin{array}{l}
			tra{n_{ijj'}} = \max \left\{ {{\Gamma _{fre}}\left( {fr{e_{ij}},fr{e_{ij'}}} \right),0,\Delta } \right.
			{\Gamma _{band}}\left( {ban{d_{ij}},ban{d_{ij'}}} \right),{\Gamma _{pol}}\left( {po{l_{ij}},po{l_{ij'}}} \right)\\
			\left. {{\Gamma _{mode}}\left( {mod{e_{ij}},mod{e_{ij'}}} \right)} \right\}
		\end{array}
	\end{equation}
	
	where ${\Gamma _{pol}}$ is the satellite polarization transition time, 
	${\Gamma _{mode}}$ is the satellite detection mode transition time, ${\Gamma _{band}}$ is the satellite bandwidth setting transition time, ${\Gamma _{fre}}$ is the satellite frequency transition time,  $\Delta $ is the satellite load on/off time.
	
	\textcolor[rgb]{0,0,0}{The scheduling goal of the EDSSP problem is to obtain the highest detection profit.} The objective function is represented as follows.
	
	\textbf{Objective function:}
	
	\begin{equation}\max \sum\limits_{i \in S} {\sum\limits_{j \in T} {\sum\limits_{o \in {O_i}} {\sum\limits_{k \in TW} {{G_j}} } } }  \cdot \Omega \left[ {\varphi \left( {degre{e_j}} \right)} \right] \cdot {x_{ijok}}
		\label{objective}
	\end{equation}
	
	where ${G_j}$ is signal gain can be obtained from the task, $\Omega \left[ {\varphi \left( {degre{e_j}} \right)} \right]$ is the gain due to the bandwidth setting, \textcolor[rgb]{0,0,0}{the product of ${G_j}$ and $\Omega \left[ {\varphi \left( {degre{e_j}} \right)} \right]$ represents the profit of the task.}
	
	\textbf{Constraints:}
	
	\begin{equation}s{t_{ijo}} \ge es{t_j} \cdot {x_{ijok}},\forall i \in S,j \in T,o \in {O_i},k \in TW\end{equation}
	\begin{equation}\left( {s{t_{ijo}} + du{r_j}} \right) \cdot {x_{ijok}} \le le{t_j},\forall i \in S,j \in T,o \in {O_i},k \in TW\end{equation}
	\begin{equation}\theta _{_{ijok}}^t \le \theta _j^{\max },\forall i \in S,j \in T,o \in {O_i},k \in TW,t \in \left[ {s{t_{ijo}},s{t_{ijo}} + du{r_j}} \right]
		\label{angle}
	\end{equation}
	\begin{equation}s{t_{ijo}}  \ge EV{T_{ijok}} \cdot {x_{ijok}},\forall i \in S,j \in T,o \in {O_i},k \in TW\end{equation}
	\begin{equation}\begin{array}{l}
			\left( {s{t_{ijo}} + du{r_j}} \right) \cdot {x_{ijok}} \le LV{T_{ijok}} ,\forall i \in S,j \in T,o \in {O_i},k \in TW
		\end{array}
		\label{tw time}
	\end{equation}
	\begin{equation}\begin{array}{*{20}{c}}
			{\sum\limits_{i \in S} {\sum\limits_{j \in T\backslash \left\{ {j'} \right\}} {\sum\limits_{k \in TW} {{m_j} \cdot {x_{ijok}}} } }  + {m_{j'}} \cdot {x_{ij'ok'}} \le M} {,\forall i \in S,j \in T,o \in {O_i},k \in TW}
	\end{array}\end{equation}
	\begin{equation}\begin{array}{*{20}{c}}
			{\left( {s{t_{ijo}} + du{r_j}} \right) \cdot {x_{ijok}} + tra{n_{ijj'}} \le s{t_{ij'o}} + I \cdot \left( {1 - {x_{ij'ok'}}} \right)} {,j \ne j',i \in S,j \in T,o \in {O_i},k \in TW}
	\end{array}\end{equation}
	\begin{equation}\sum\limits_{i \in S} {\sum\limits_{o \in {O_i}} {\sum\limits_{k \in TW} {{x_{ijok}}} } }  \le 1,\forall i \in S,o \in {O_i},k \in TW\end{equation}
	\begin{equation}{x_{ijok}} \in \left\{ {0,1} \right\}\end{equation}
	
	\textcolor[rgb]{0,0,0}{Constraints 1-2 indicate that the start time and end time of the task must be within the time range required. Constraint 3 indicates that the detection angle cannot exceed the task maximum angle requirement. Constraints 4-5 indicate that the start time and end time are within the visible time window. Constraint 6 indicates that the satellite cannot exceed the upper limit of the satellite storage capacity in each orbit. Constraint 7 indicates that the satellite must meet various transition time requirements to perform every two tasks. Constraint 8 indicates that each task can be executed at most once. Constraint 9 indicates the value range of the decision variable.}
	
	\section{The Proposed Method}
	Since the single satellite imaging scheduling problem has been proved to be NP-hard,  the more complex EDSSP problem does not exist polynomial time algorithm\cite{b33,b34}. Therefore, we propose a reinforcement learning-based genetic algorithm to solve the EDSSP problem. In addition, we also design a task time window selection algorithm for generating schemes. The reinforcement learning-based genetic algorithm, details in evolutionary algorithms, and strategies in reinforcement learning and task time window selection algorithms are given in this section.

\subsection{Reinforcement Learning based Genetic Algorithm (RL-GA)}
\begin{algorithm}[htp]
	\caption{Reinforcement Learning based Genetic Algorithm (RL-GA)}
	\LinesNumbered
	\KwIn{population size ${N_{p}}$, learning rate $\alpha  $, discount factor $\gamma $, Q-table $Q $, $\varepsilon $, control parameter $T$, crossover operator ${C_o}$, mutation operator ${M_o}$, $L$}
	\KwOut{${Solution}$}
	Initialize algorithm parameters and the population\;
	Set $t=1$, $num\_eval=0$, $local\_best=0$, $count=0$,  $global\_best=0$, $global\_best_individual=\left[ \ \right]$, $local\_best\_individual=\left[ \ \right]$\;
	Initialize Q-table${(C_o, M_o)}$\;
	\While{$num\_eval < MFE$}{
		\For{$p=1$ to ${N_{p}}$}{
			$\rho \left( {{s_i},{a_j}} \right) \leftarrow $Use softmax strategy$\left( {{Q_t}\left( {{s_i},{a_j}} \right),T} \right)$\;
			$A_p^t \leftarrow $Select action with Q-Learning$\left(\varepsilon, \rho \left( {{s_i},{a_j}}, T \right)\right)$\;
			${indi_p^{t-1}} \leftarrow$Roulette selection of individual $\left({P_{t-1}},{F_{t-1}}\right)$\;
			${indi_p^{t}} \leftarrow$Evolution Operation$\left({indi_p^{t-1}}, A_p^t \right)$// A combination of crossover and mutation operators\;
			${S_p^{t}} \leftarrow$Generate a plan$\left(indi_p^{t},T,TW\right)$ //Use Task Time Window Selection Algorithm\;
			${F_p^{t}} \leftarrow$Fitness Evaluation$\left({S_p^{t}}\right)$ by Eq. \ref{objective}\;
			${R_{t}}\leftarrow$Compute Reward$\left({F_{t}}, {F_{t-1}}\right)$ by Eq. \ref{reward eq}\;
			${Q_{t+1}}\leftarrow$Compute Q-values$\left( {{S_{t}},{A_{t}}},{R_{t}},\alpha,\gamma \right)$\;
			$t\leftarrow t + 1$\;
			${num\_eval} \leftarrow num\_eval+1$\;
		}
		
		${local\_best} \leftarrow$Find the max fitness of current population$\left(P_t\right)$\;
		\eIf{$local\_best > global\_best$}{
			$global\_best\leftarrow local\_best$\;
			$global\_best\_individual\leftarrow local\_best\_individual$\;}{
			\If{$count$ not equal to $Thre$}{
				//Retain elite individuals\\
				$P_t\leftarrow$ Replace the individual of $selected\_id$ with $gobal\_best\_individual$\;
			}
		}
		\If{$local\_best <= last\_local\_best$}{
			$count\leftarrow count+1$\;
		}
		$last\_local\_best\leftarrow local\_best$\;
}\end{algorithm}

Evolutionary algorithms simulate the evolution process of biological populations and find high-quality solutions to problems in the form of population evolution. Population search brings a strong global search capability to evolutionary algorithms, which ensures its performance in solving large-scale complex problems.

The obvious disadvantage of evolutionary algorithms is that the search results are extremely sensitive to the parameter configuration. This feature makes many evolutionary algorithms have dependencies on specific problems and scenarios. When problems or scenarios change, parameters need to be adjusted or reset. The parameter tuning process is time-consuming. According to the characteristics of the evolutionary algorithm and reinforcement learning method, we design a reinforcement learning-based genetic algorithm for the EDSSP problem. Genetic algorithm has good global search ability, and it has a good performance in the field of sequence optimization and scheduling, such as traveling salesman problems, vehicle route planning problems, satellite task scheduling problems, and many other problems. To improve the local search ability of the genetic algorithm, we introduce the elite individual retention strategy into the genetic algorithm. RL-GA uses a Q-learning method to complete the evolution of the population. The Q-learning method is responsible for selecting crossover and mutation operations. The population obtained by the RL-GA algorithm uses a task time window selection algorithm (TTWSA) to complete the decoding and generate the task execution plan. The profit obtained by the TTWSA is used for the reward calculation and then the Q value update process. In other words, TTWSA provides a basis for the learning method to select new actions.
The pseudo-code of the genetic algorithm based on Q-learning is shown in Algorithm Table 2.

\begin{center}
	\begin{figure}[htp]
		\centerline{\includegraphics[scale=0.5,trim=0 0 0 0]{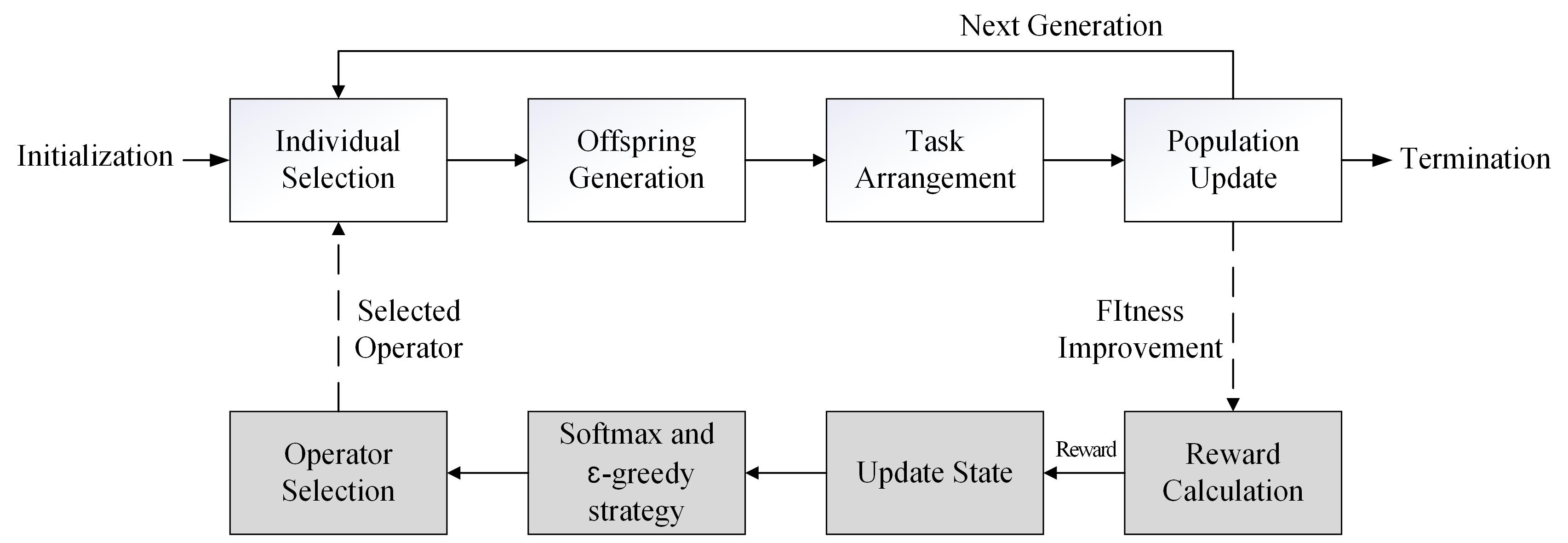}}
		\caption{Dataflow between the RL method and GA}
		\label{pic dataflow}
	\end{figure}
\end{center}

In RL-GA, $count$ is a parameter to control whether the algorithm executes elite individual retention. When $count$ is equal to $Thre$ (Line 22), elite individuals are no longer retained. Obtaining the task planning scheme from the genetic algorithm needs to use the task time window selection algorithm. The TTWSA can determine the tasks that can be executed and determine the specific start time and end time. \textcolor[rgb]{0,0,0}{Figure \ref{pic dataflow} shows the dataflow between the reinforcement learning method and the genetic algorithm population search.}

\textcolor[rgb]{0,0,0}{As shown in Figure \ref{pic dataflow}, the top row of the figure represents the GA population evolution process, and the bottom row represents the Q-learning algorithm choosing the operator, updating the state, etc. It can be seen from the figure that there is a close connection between GA and Q-learning method. The GA provides data input for the agent to select actions, and the actions selected by the agent are used directly in the population evolution.}

\subsubsection{Individual Representation and Initialization}
\textcolor[rgb]{0,0,0}{The initialization method is the first step in the population search to obtain a detection scheme. It needs to construct an initial solution in a certain way and further search to find a higher-quality solution based on the initial solution. In the RL-GA, individuals are encoded in real numbers. For decoding, the RL-GA uses the task time window selection algorithm to determine whether to execute each task according to the order of tasks. If one task can be executed, the specific execution time of the task will be determined. Otherwise, the task will be discarded. Eventually, the scheme will be generated by arranging the tasks one by one. The advantage of real number coding is to ensure the unique correspondence between the elements at each position in the sequence, and the decoding process is simple. Another benefit of this individual representation method is that it effectively guarantees the legitimacy of the solution structure without using any repair methods.}

\subsubsection{Fitness Evaluation}
\textcolor[rgb]{0,0,0}{Fitness evaluation method has two roles in the RL-GA algorithm, one is to select individuals from the population, and the other is to evaluate the reward and update state in the reinforcement learning method.} The fitness evaluation is obtained by calculating the objective function value of each corresponding to the plan by Eq. \ref{objective}. The method for generating plans from individuals is introduced in section 3.3.

\subsubsection{Individual Selection}
Individual selection is the premise of population evolution. After an individual is selected, the corresponding operation chosen by the reinforcement learning method can be completed to obtain a new individual. On the one hand, individual selection should make good performers more likely to be successfully selected. On the other hand, other individuals should also have a certain possibility of being selected to ensure the diversity of the population. \textcolor[rgb]{0,0,0}{Therefore, we use a roulette approach to select an individual in the RL-GA. The calculation formula for selecting individuals by roulette is as follows.}

\begin{equation}{p_i} = \frac{{fi{t_i}}}{{\sum\limits_{i = 1}^{N_p} {fi{t_i}} }}\end{equation}

where ${fi{t_i}}$ represents the fitness function value of individual. The fitness function value is calculated using Eq. \ref{objective}.
\subsubsection{Crossover}
\textcolor[rgb]{0,0,0}{The purpose of evolution is to improve the performance of individuals within a population. We use crossover and mutation operators to search for good detection plans. For the crossover operation in the RL-GA, we design seven crossover operators with various lengths and heuristic rules, which are called short segment crossover operator (denoted as C1), medium segment crossover operator (denoted as C2), and long segment crossover operator (denoted as C3), segment flip (denoted as C4), foremost position crossover (denoted as C5), segment internal order adjustment based on earliest available start time (denoted as C6),  fragment internal order adjustment based on task duration (denoted as C7).}

\begin{center}
	\begin{figure}[htp]
		\centerline{\includegraphics[scale=1,trim=0 0 0 0]{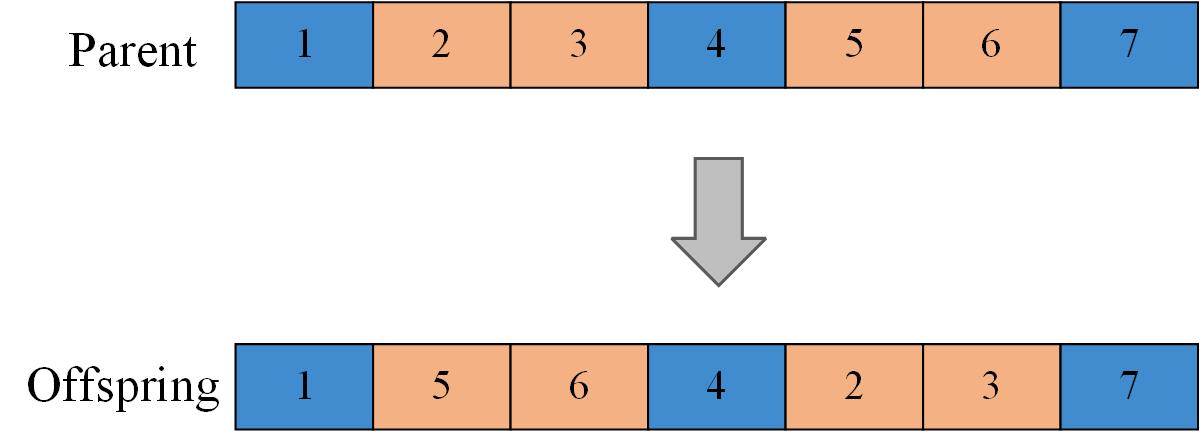}}
		\caption{Schematic Diagram of the Short Segment Crossover }
		\label{pic crossover}
	\end{figure}
\end{center}

\textbf{Short segment crossover operator (C1): }Short segment crossover refers to selecting two task sequence segments of length $L$ from an individual and exchanging the positions of the two segments in the individual. The schematic diagram of the short segment crossover is shown in Figure \ref{pic crossover}.

\textbf{Medium segment crossover operator (C2):} Medium segment crossover refers to selecting two task sequence segments of length $2L$ from an individual and swapping the positions of the two segments in the individual.

\textbf{Long segment crossover operator (C3): }Long segment crossover refers to selecting two task sequence segments of length $3L$ from an individual and exchanging the positions of the two segments in the individual.

\begin{center}
	\begin{figure}[htp]
		\centerline{\includegraphics[scale=1,trim=0 0 0 0]{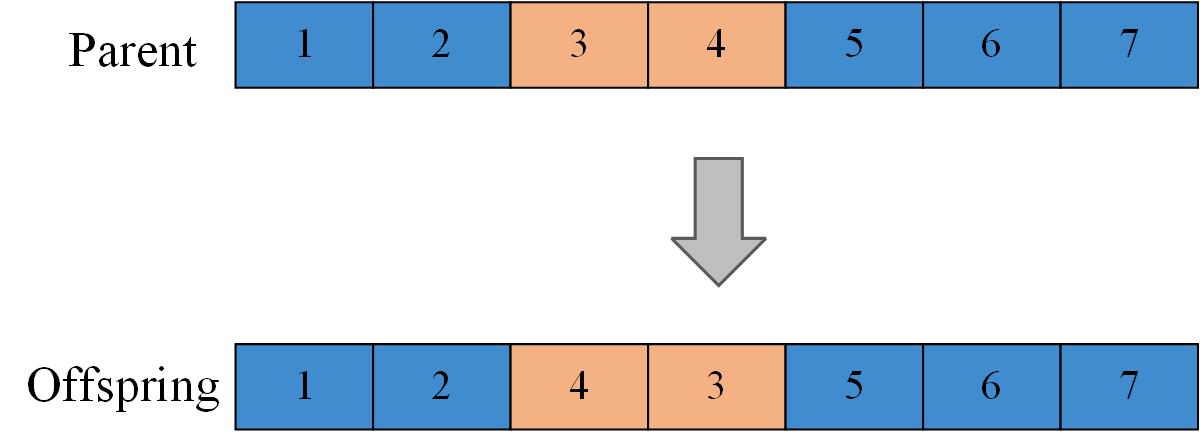}}
		\caption{Schematic Diagram of the  Fragment Flipping }
		\label{pic flip}
	\end{figure}
\end{center}

\textbf{Fragment flipping (C4): }Fragment flipping is to select a task sequence fragment of length $L$ from an individual and rearrange the task sequence from back to front. The schematic diagram of the fragment flipping is shown in Figure \ref{pic flip}.

\textbf{Foremost position crossover (C5): }Select a task sequence segment of length $L$ from an individual, and swap the segment with the task sequence segment of the same length at the foremost position of the individual.

\textbf{Segment internal order adjustment based on earliest available start time (C6): }Select a task segment of length $L$ from an individual and adjust the internal segment order from front to back according to its task's earliest available start time $est_j$.

\textbf{Segment internal order adjustment based on task duration (C7): }Select a task segment of length $L$ from an individual and adjust the internal segment sequence from short to long according to the task required duration $dur_j$.

\subsubsection{Mutation}
\textcolor[rgb]{0,0,0}{Compared with crossover, the variation degree of mutation is smaller, which is also helpful for improving the search performance. The mutation in RL-GA is performed by exchanging two positions in the sequence. We combine the crossover operator and mutation operator to complete the individual evolution of the population. After the evolution process is completed, the fitness of offspring needs to be evaluated. The TTWSA is used to generate a plan and the fitness value of each individual is calculated by Eq. \ref{objective}. In the RL-GA, the crossover and mutation operators used for optimization are chosen by the Q-learning method, and the operation used for individual evolution is selected according to the Q value. }

\subsubsection{Reinforcement Learning Method}

We adopt the Q-learning method to guide the search for an evolutionary algorithm. Q-learning is a simple and effective reinforcement learning method, which is widely used in various models such as DQN and DRL according to the actions taken by the algorithm adaptive decision-making according to the Q value\cite{b35,b36}. Q-learning consists of five parts: action, state, reward, learning algorithm, and environment, which can be represented as a quadruple of $\left\langle {A,S,R,L,E} \right\rangle $\cite{b37}. \textcolor[rgb]{0,0,0}{The state of the agent indicates whether the fitness function value has improved through the search, and is divided into two states, one with improvement and the other without improvement or reduction. Such a state representation can be associated with the improvement of the population fitness value or can be combined with optional actions to form a Q-table. In this way, the Q-table is simple and can be closely integrated with the population evolution process. The actions represent the combination of population search operators used, consisting of several crossover and mutation operators. }

\begin{center}
	\begin{figure}[htp]
		\centerline{\includegraphics[scale=0.7,trim=0 0 0 0]{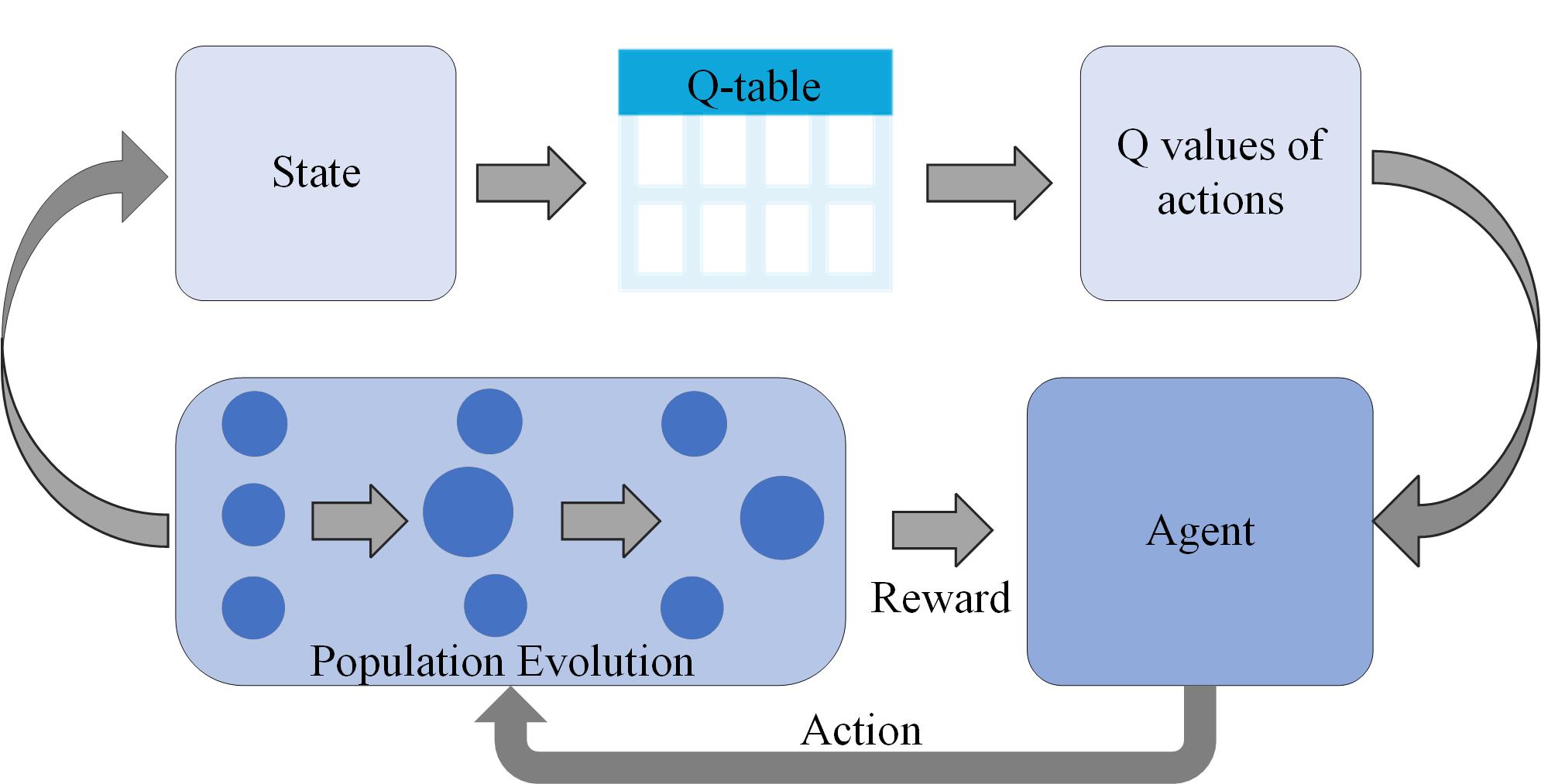}}
		\caption{Q-learning Method Decision-making Process for EDSSP}
		\label{pic Qlearning}
	\end{figure}
\end{center}

The Q-learning method selects the population update strategy through the update of the Q value, which replaces the traditional operator selection method in the evolutionary algorithm. More specifically, the agent selects the appropriate operator for population evolution based on the Q-value and action selection strategy. The interaction between the agent and the environment is evaluated by the reward function. The reward value is closely related to the fitness value and fitness improvement. The running process of Q-learning is shown in Figure \ref{pic Qlearning}.

\textcolor[rgb]{0,0,0}{As shown in Figure \ref{pic Qlearning}, the RL-GA algorithm generates offspring after each population evolution search using the operators decided by the Q-learning method. And the task time window selection algorithm is used to generate detection plans. The value of the fitness function can be easily calculated based on the plan. The Q-learning method can further obtain fitness improvement. The state will update according to the fitness improvement.}

The formula for calculating reward is shown below.
\begin{equation}{R_{t}} = {F_{t}}\left( {{S_{t}},A_t} \right) - {F_{t-1}}\left( {{S_{t-1}},A_{t-1}} \right)
	\label{reward eq}
\end{equation}

where ${F_{t}}$ and ${F_{t-1}}$ are values of fitness at the time $t$ and $t-1$, respectively.

The updated formula for the Q value is shown below.
\begin{equation}\begin{array}{l}
		{Q_{t + 1}}\left( {{S_t},{A_t}} \right) = {Q_t}\left( {{S_t},{A_t}} \right)
		+ \alpha \left( {{R_{t}} + \gamma \mathop {\max }\limits_a {\rm{ }}Q\left( {{S_{t + 1}},a} \right) - {Q}\left( {{S_t},{A_t}} \right)} \right)
\end{array}\end{equation}

where $\alpha$ is the learning rate, and $\gamma$ is the discount factor.

An important part of Q-learning is the Q-value table, which records the performance values of all crossover and mutation combinations used in the search process. When the RL-GA algorithm contains ${n_c}$ kinds of crossover operators and ${n_m}$ kinds of mutation operators, the total number of actions contained in the Q-value table is ${n_c} \times {n_m} + {n_c} + {n_m}$. In our proposed algorithm, there are 7 kinds of crossover operators and 1 kind of mutation operator. Therefore, the number of actions that each agent can choose is 15. We set the state to two levels according to the fitness function value improvement, which is higher (denoted as I) and lower (denoted as II). In the proposed algorithm, the number of $<$state, action$>$ combinations is 30.
For each agent, the probability of selecting $a_j$ action in state $s_i$ is usually determined by Softmax strategy:
\begin{equation}\rho \left( {{s_i},{a_j}} \right) = \frac{{{e^{{{{Q_t}\left( {{s_i},{a_j}} \right)} \mathord{\left/
							{\vphantom {{{Q_t}\left( {{s_i},{a_j}} \right)} T}} \right.
							\kern-\nulldelimiterspace} T}}}}}{{\sum\nolimits_{j = 1}^n {{e^{{{{Q_t}\left( {{s_i},{a_j}} \right)} \mathord{\left/
								{\vphantom {{{Q_t}\left( {{s_i},{a_j}} \right)} T}} \right.
								\kern-\nulldelimiterspace} T}}}} }}\end{equation}

where ${{Q_t}\left( {{s_i},{a_j}} \right)}$ is the corresponding value in the Q table at time $t$, and $T$ is a control parameter.

Due to the large value of the objective function in the EDSSP problem, the introduction of a control parameter $T$ can effectively prevent the probability value of individuals from being too large. Meanwhile, the reinforcement learning method needs to balance the exploration and exploitation of algorithm\cite{b10}. To prevent the algorithm from falling into a local optimum that is difficult to jump out, we use a parameter $\varepsilon $ to regulate this process. When the randomly generated probability value is less than $\varepsilon $, a random evolution strategy is selected to generate offspring.

\subsubsection{Elite Individual Retention Strategies}
\textcolor[rgb]{0,0,0}{Since the EDSSP problem has a large solution space, we use an elite individual retention strategy in the RL-GA to improve the convergence performance.} When the retention process occurs, the optimal individual obtained by the search is retained in the next generation population. The use of elite individuals will affect the global search ability of the algorithm after a certain period of search. Therefore, we introduce a parameter $count$ and a threshold $Thre$. When the current population search is not improved compared to the previous generation population, i.e., the maximum fitness value of the population does not change. The parameter $count$ will increase by 1. If the value of parameter $count$ is equal to the threshold $Thre$, the elite individual is no longer retained.

\subsubsection{Termination} 
\textcolor[rgb]{0,0,0}{After a certain number of iterative searches, the RL-GA needs to finish the search process and output the optimal result as the final execution plan. Since we adopt the crossover and mutation operators based on reinforcement learning, when the agent chooses an action, it can correspond to one or two evolution operations. Therefore, it is reasonable to use the evolution times $num\_eval$ as the evaluation index to record the iterative search.} When the algorithm evolution times reach the set maximum evolution times $MFE$, the algorithm search process will be terminated.

\subsection{Task Time Window Selection Algorithm (TTWSA)}

When the task execution is determined through algorithm optimization, the start time and end time of the task need to be determined. \textcolor[rgb]{0,0,0}{The TTWSA algorithm tries to schedule the tasks in the best position to obtain the maximum detection profit and adjusts the execution time of the tasks according to the actual situation faced. The heuristic rules used in the algorithm refer to the approach used in \cite{b31,b32} and are designed accordingly to the EDSSP problem.} The setting of task start time and end time not only affects the detection profit of the task but also affects the satisfaction of constraints. We propose a two-stage task time window selection algorithm with preliminary filtering and then determine the execution time.
The purpose of preliminary filtering is to ensure that the task execution can meet the requirements of constraints, such as angle, transition time, and others. Therefore, the preliminary filtering needs to clip some time windows that do not meet the constraints or discard the current time window and try to schedule tasks in another time window. The basis of preliminary filtering is to obtain the actual earliest actual available start time ${AEV{T_{ijok}}}$ and the latest actual available end time ${ALV{T_{ijok}}}$. The preliminary filtering is based on the Eq. \ref{angle}-\ref{tw time}. The formulas of ${AEV{T_{ijok}}}$  and ${ALV{T_{ijok}}}$ are shown as Eq. \ref{eqaev} and Eq. \ref{eqalv}  respectively.
\begin{equation}
	AEV{T_{ijok}} = \max \left\{ {es{t_j},EV{T_{ijok}}} \right\}\label{eqaev}
\end{equation}
\begin{equation}
	ALV{T_{ijok}} = \min \left\{ {le{t_j},LV{T_{ijok}}} \right\}\label{eqalv}
\end{equation}

After the preliminary filtering, the algorithm can specify the start time and end time of the task. \textcolor[rgb]{0,0,0}{Detection profit that can be obtained follows the rule of first growing and then decreasing.} To obtain more profit, the detection process should ensure that the angle between the antenna and the signal source is as small as possible. The time range in which the algorithm chooses to execute the task should allow for sufficient profit. Generally speaking, the change in the angle between the satellite and the target in the time window can be regarded as a symmetrical process. Therefore, the middle position of the time window is the best choice as an intermediate moment, which usually has the smallest angle. And the start execution time and the task end time are determined according to this intermediate moment. The formula for calculating the best start time is as follows.
\begin{equation}bst_{ijo}=\left(EVT_{ijok}+LVT_{ijok}\right) / 2-dur_j/ 2\end{equation}
\begin{equation}
	be{t_{ijo}} = bs{t_{ijo}} + du{r_j}
	\label{best end time}
\end{equation}

where $bs{t_{ijo}}$ represents the best start time of the execution task $j$ of satellite $i$ on orbit $o$.

If this moment can meet the requirements of other constraints, it will be regarded as the final start time; otherwise, move the execution time backward or forward until all constraints are met. The pseudo-code of the task time window selection algorithm is shown in the Algorithm Table \ref{task time window}.

\begin{algorithm}[ht]
	\label{task time window}
	\caption{Task Time Window Selection Algorithm}
	\LinesNumbered 
	\KwIn{ Task Set ${T}$, Time Window ${TW}$, Detection Task Sequence ${indi_p^t}$}
	\KwOut{Solution $S_i^t$}
	\ForEach{task ${t_i}$ in ${T}$ in the order in ${indi_p^t}$}
	{
		\ForEach{time window ${tw_j}$ in ${TW}$}
		{
			${AEV{T_{ijok}} = \max \left\{ {es{t_j},EV{T_{ijok}}} \right\}}$ \;
			${ALV{T_{ijok}} = \min \left\{ {le{t_j},LV{T_{ijok}}} \right\}}$ \;
			\eIf{${(ALV{T_{ijok}}-AEV{T_{ijok}}) \ge d}$ }
			{
				$bs{t_{ijo}} \leftarrow $Calculate the best start time by Eq. \ref{best start time}\;
				$be{t_{ijo}} \leftarrow $Calculate the best end time by Eq. \ref{best end time}\;
				\If{$bs{t_{ijo}}<AEV{T_{ijok}}$}
				{
					$s{t_{ijo}} \leftarrow $ move backward to determine start time\;
				}
				\If{$be{t_{ijo}}>ALV{T_{ijok}}$}
				{
					$s{t_{ijo}} \leftarrow $ move forward to determine start time\;
				}
				$t{w'_j},t{w''_j} \leftarrow $Update ${tw_j}$\;
				$TW \leftarrow $ Add $t{w'_j}$ and $t{w''_j}$ into $TW$\;
				$TW \leftarrow $Omit ${tw_j}$ from $TW$\;
				Try to arrange next task ${t_{i+1}}$\;
			}
			{
				Turn to next time window $tw_{j+1}$\;
			}
		}
	}
\end{algorithm}

\section{Performance Evaluation of Proposed Algorithm}

\subsection{Experimental Settings}

The configuration of the experiment in this paper is Core I7-7700 3.6 GHz CPU, 16 GB memory, Windows 11 operating system desktop computer, and Matlab 2020a is used for coding. All algorithms are run under the same system configuration.

We use a series of Chinese satellites for our experiments. The satellite orbit parameters can be defined by a hexagram: the length of semi-major axis $\left( a \right)$, eccentricity $\left( e \right)$, inclination $\left( i \right)$, argument of perigee $\left( \omega  \right)$, right ascension of the ascending node $\left( \varphi  \right)$, and mean anomaly $\left( m \right)$. Table 1 gives the initial orbital parameters of two satellites.

\begin{table}[ht]
	\caption{Initial Orbital Parameters of Two Satellites}
	\centering
	\begin{tabular}{ccccccc}
		\toprule
		Satellite & $a$ & $e$ & $i$ & $\omega$ & $\varphi$ & $m$ \\
		\midrule
		No. 1 & 7000 & 0.00015 & 97.672 & 0 & 21.75 & 158.25 \\
		No. 2 & 7000 & 0.00015 & 97.672 & 0 & 51.75 & 128.25 \\
		… & … & … & … & … & … & …\\
		\bottomrule
	\end{tabular}
\end{table}

\textcolor[rgb]{0,0,0}{We use AGI Systems Tool Kit (STK) 11.2 to generate the task and time window data needed for scheduling. The task distribution is randomly generated on a global scale. The planning time horizon of the tasks is all within one day. For a detection task, the duration follows a uniform distribution in the interval $\left[10,100\right]$ seconds, and the mission requires a maximum time horizon of 12 hours to execute. Each task is completed with only one detection activity by satellite. When the bandwidth used by the satellite is set to $bandwidth_1$, $bandwidth_2$, $bandwidth_3$, $bandwidth_4$, and $bandwidth_5$, the signal gains obtained obey a uniform distribution in the intervals $\left[1,3\right]$, $\left[4,6\right]$, $\left[7,9\right]$, $\left[10,12\right]$, and $\left[13,15\right]$, respectively.}

To verify the effect of our proposed algorithm. We choose an improved genetic algorithm (IGA)\cite{b38}, an adaptive large neighborhood search algorithm (ALNS/TPF)\cite{b39}, an artificial bee colony algorithm (ABC) \cite{b40}, and a construction heuristic algorithm (CHA) as state-of-the-art algorithms. The IGA algorithm adopts a heuristic mechanism and uses knowledge to guide the algorithm search. The ALNS/TPF algorithm combines tabu search and adaptive large neighborhood search algorithm to find high-quality solutions through destruction and repair. The ABC algorithm search solution space (also known as nectar sources) through three types of bees, imitating the way bee populations search. The CHA algorithm gives the planned task sequence according to the way of sorting the task profit. The above four algorithms have a large number of applications in satellite task scheduling problems and other planning problems and have good performance.

The evaluation times $MFE$ of all algorithms were set to 5000 times. The control parameter $T$ of RL-GA is set to 1000, the initial learning rate $\alpha$ is set to 0.01, the discount factor $\gamma$ is set to 0.95, the threshold $Thre$ is set to 100, the length $L$ is 2 and $\varepsilon$ is 0.01. The crossover probability of the IGA algorithm was set to 0.9, and the mutation probability was set to 0.1. The initialized score of ALNS/TPF is 100, and the increased scores according to performance are 50, 20, and 10, respectively, and the tabu list length is 10. The population size $N_p$ of RL-GA, IGA, and ABC are all set to 10.

We design four scale instances, named small-scale (100-300 tasks), medium-scale (400-600 tasks), large-scale (700-800 tasks), and ultra-large-scale (1000-1400 tasks). Each instance is marked in the form of "A-B", "A" represents the number of tasks, and "B" represents the internal number of the instance under that number. Detection tasks are randomly distributed around the world. Each algorithm was run 30 times on an instance. The detection profit of the task sequence is the basis for the algorithm evaluation. For the evaluation benchmark metrics, we separately count the maximum (denoted as Best), mean (denoted as Mean), and standard deviation (denoted as Std. Dev.) of the obtained results.
\textcolor[rgb]{0,0,0}{We also use the Wilcoxon rank sum test to determine whether there is a significant difference between the search results of different algorithms, and the significance level is chosen as $p=0.05$}. In addition, the CPU time and convergence speed of the algorithm are also used to evaluate algorithms.

\subsection{Results}

\subsubsection{Evaluation of scheduling performance}
\begin{table}[!htp]
	\tiny
	\caption{Scheduling Results for Small-scale Instances}
	\centering
	\label{table for small scale}
	\begin{tabular}{cccccccccc}
		\toprule
		\multirow{2}{*}{Instance} & \multicolumn{2}{l}{RL-GA} & \multicolumn{2}{l}{IGA} & \multicolumn{2}{l}{ALNS/TPF} & \multicolumn{2}{l}{ABC} & \multirow{2}{*}{CHA} \\
		\cmidrule(r){2-3}  \cmidrule(r){4-5} \cmidrule(r){6-7} \cmidrule(r){8-9}
		& Best & Mean(Std. Dev.) & Best & Mean(Std. Dev.) & Best & Mean(Std. Dev.) & Best & Mean(Std. Dev.) &  \\
		\midrule
		100-1 & \textbf{823} & 820(1.19) & \textbf{823} & 819.6(1.55)= & \textbf{823} & 817.5(2.89)= & \textbf{823} & 819.07(1.46)= & 790 \\
		100-2 & \textbf{801} & 801(0) & \textbf{801} & 801(0)= & \textbf{801} & 801(0)= & \textbf{801} & 801(0)= & 801 \\
		100-3 & \textbf{813} & 813(0) & \textbf{813} & 813(0)= & \textbf{813} & 813(0)= & \textbf{813} & 813(0)= & 813 \\
		100-4 & \textbf{787} & 787(0) & \textbf{787} & 787(0)= & \textbf{787} & 787(0)= & \textbf{787} & 787(0)= & 787 \\
		200-1 & \textbf{1657} & 1629.07(5.82) & 1617 & 1604.77(12.62)- & 1616 & 1601.77(8.34)- & 1618 & 1607.33(5.93)= & 1513 \\
		200-2 & \textbf{1499} & 1493.4(3.23) & 1494 & 1487.6(3.81)- & 1490 & 1484.87(3.29)- & 1498 & 1489.6(3.62)= & 1406 \\
		200-3 & \textbf{1615} & 1603.5(4.81) & 1599 & 1589.5(6.53)- & 1604 & 1588.5(5.15)- & 1603 & 1592.7(5.12)- & 1456 \\
		200-4 & \textbf{1732} & 1713.17(6.26) & 1702 & 1691.3(9.83)- & 1711 & 1692.27(8.19)- & 1706 & 1694.17(6.45)- & 1598 \\
		300-1 & \textbf{2180} & 2149.67(8.36) & 2091 & 2071.97(16.73)- & 2086 & 2066.93(9.47)- & 2111 & 2080.4(11.4)- & 1869 \\
		300-2 & \textbf{2211} & 2159.6(9.75) & 2111 & 2078.83(17.14)- & 2107 & 2078.5(13.83)- & 2111 & 2090.4(12.3)- & 1811 \\
		300-3 & \textbf{2145} & 2111.5(10.32) & 2057 & 2033.13(23.22)- & 2052 & 2033.9(11.16)- & 2062 & 2042.67(11.03)- & 1773 \\
		300-4 & \textbf{2160} & 2122.47(8.94) & 2087 & 2065(15.27)- & 2098 & 2066.2(13.3)- & 2112 & 2074.77(11.94)- & 1826\\
		\bottomrule
	\end{tabular}
\end{table}
The results of the small-scale instances are shown in Table 2. From the results, it can be seen that the profit gap between the proposed algorithm and the state-of-the-art algorithm increases with the increase of the problem scale. There is no difference in the results between all five algorithms except for the maximum profit for 100 tasks. When the task scale increases to 200, the gap between the algorithms becomes significant. It is worth noting that when the task scale is increased to 300, the maximum benefit that can be obtained from the search does not increase in multiples of the task scale. This means that the influence of constraints on task scheduling increases, and some tasks that violate constraints cannot be executed.

\begin{table}[!htp]
	\tiny
	\caption{Scheduling Results for Medium-scale and Large-scale Instances}
	\centering
	\label{table for mid and large scale}
	\begin{tabular}{cccccccccc}
		\toprule
		\multirow{2}{*}{Instance} & \multicolumn{2}{l}{RL-GA} & \multicolumn{2}{l}{IGA} & \multicolumn{2}{l}{ALNS/TPF} & \multicolumn{2}{l}{ABC} & \multirow{2}{*}{CHA} \\
		\cmidrule(r){2-3}  \cmidrule(r){4-5} \cmidrule(r){6-7} \cmidrule(r){8-9}
		& Best & Mean(Std. Dev.) & Best & Mean(Std. Dev.) & Best & Mean(Std. Dev.) & Best & Mean(Std. Dev.) &  \\
		\midrule
		400-1 & \textbf{3130} & 3110.8(5.96) & 3092 & 3079.17(6.31)- & 3097 & 3080.47(7.7)- & 3103 & 3087.87(6.64)- & 2906 \\
		400-2 & \textbf{3052} & 3035.17(5.59) & 3017 & 3000.4(6.93)- & 3014 & 2993.47(8.15)- & 3016 & 3003.07(6.27)- & 2881 \\
		400-3 & \textbf{3069} & 3057.93(5.53) & 3056 & 3045.17(5.83)- & 3058 & 3046.37(5.79)- & 3058 & 3047.97(3.93)= & 2917 \\
		400-4 & \textbf{3223} & 3215.57(6.77) & 3211 & 3196.97(6.77)- & 3212 & 3196.9(8.16)- & 3217 & 3201.83(6.4)- & 3101 \\
		500-1 & \textbf{3985} & 3953.67(8.28) & 3928 & 3893.93(11.1)- & 3911 & 3888.43(10.97)- & 3921 & 3901.53(9.27)- & 3651 \\
		500-2 & \textbf{3702} & 3663.6(10.12) & 3625 & 3592.23(11.56)- & 3632 & 3592.13(14.09)- & 3627 & 3598.83(12.61)- & 3322 \\
		500-3 & \textbf{3635} & 3598.27(10.76) & 3540 & 3508.93(11.49)- & 3548 & 3509.43(16.05)- & 3547 & 3522.07(12.43)- & 3284 \\
		500-4 & \textbf{3932} & 3910.63(10.52) & 3851 & 3831.97(11.5)- & 3881 & 3827.7(15.6)- & 3866 & 3837.53(14.56)- & 3539 \\
		600-1 & \textbf{4367} & 4303.67(11.85) & 4224 & 4191.93(15.74)- & 4214 & 4186.63(14.98)- & 4237 & 4208.67(10.65)- & 3758 \\
		600-2 & \textbf{4414} & 4369.47(10.72) & 4297 & 4237(19.41)- & 4297 & 4233.33(18.75)- & 4276 & 4250.5(13.06)- & 3847 \\
		600-3 & \textbf{4378} & 4328.2(9.35) & 4234 & 4212.7(11.48)- & 4244 & 4207.97(13.62)- & 4269 & 4233.87(15.65)- & 3801 \\
		600-4 & \textbf{4144} & 4099.03(12.33) & 4020 & 3987.03(14.49)- & 4012 & 3981.77(14.51)- & 4038 & 3997.97(16.02)- & 3641 \\
		700-1 & \textbf{4895} & 4801.67(16.66) & 4698 & 4651(19.89)- & 4697 & 4647.17(24.65)- & 4695 & 4665.47(19.86)- & 4133 \\
		700-2 & \textbf{4793} & 4730.93(11.28) & 4623 & 4588.5(14.47)- & 4612 & 4584.8(15.12)- & 4647 & 4605.23(16.4)- & 4061 \\
		700-3 & \textbf{4636} & 4572.43(17.76) & 4483 & 4443.03(20.48)- & 4505 & 4440.33(23.8)- & 4486 & 4457.33(13.53)- & 3974 \\
		700-4 & \textbf{4969} & 4900.33(18.54) & 4814 & 4758.07(20.54)- & 4796 & 4747.43(21.92)- & 4811 & 4768.87(18.78)- & 4334 \\
		800-1 & \textbf{4841} & 4778.2(20.59) & 4642 & 4581.8(29.58)- & 4620 & 4573.77(25.55)- & 4667 & 4613.23(23.35)- & 3815 \\
		800-2 & \textbf{4996} & 4928.03(20.85) & 4785 & 4731.8(20.94)- & 4807 & 4728.27(27.91)- & 4838 & 4769.17(27.16)- & 4126 \\
		800-3 & \textbf{5110} & 4993.37(19.8) & 4885 & 4835.5(23.71)- & 4870 & 4824.9(23.43)- & 4908 & 4859.33(21.19)- & 4171 \\
		800-4 & \textbf{5055} & 4947.07(22.68) & 4829 & 4775.5(23.46)- & 4819 & 4763.43(24.58)- & 4884 & 4801.6(24.58)- & 4153\\
		\bottomrule
	\end{tabular}
\end{table}

\begin{table}[htp]
	\tiny
	\caption{Scheduling Results for Ultra-large-scale Instances}
	\centering
	\label{table for very large scale}
	\begin{tabular}{cccccccccc}
		\toprule
		\multirow{2}{*}{Instance} & \multicolumn{2}{l}{RL-GA} & \multicolumn{2}{l}{IGA} & \multicolumn{2}{l}{ALNS/TPF} & \multicolumn{2}{l}{ABC} & \multirow{2}{*}{CHA} \\
		\cmidrule(r){2-3}  \cmidrule(r){4-5} \cmidrule(r){6-7} \cmidrule(r){8-9}
		& Best & Mean(Std. Dev.) & Best & Mean(Std. Dev.) & Best & Mean(Std. Dev.) & Best & Mean(Std. Dev.) &  \\
		\midrule
		1000-1 & \textbf{8003} & 7939.33(15.24) & 7832 & 7803.6(17.28)- & 7884 & 7806.73(24.29)- & 7886 & 7829.07(24.23)- & 7431 \\
		1000-2 & \textbf{7821} & 7788.6(16.31) & 7733 & 7691.93(16.59)- & 7726 & 7686.53(18.95)- & 7737 & 7706.93(22.45)- & 7353 \\
		1000-3 & \textbf{7907} & 7882.27(13.95) & 7852 & 7816.6(13.34)- & 7848 & 7811.9(17.65)- & 7867 & 7829.2(14.66)- & 7565 \\
		1000-4 & \textbf{8091} & 8045.57(14.01) & 7993 & 7938.73(20.21)- & 7977 & 7924.4(19.31)- & 7983 & 7947.17(18.91)- & 7616 \\
		1100-1 & \textbf{8452} & 8401.7(17.73) & 8322 & 8284.2(18.31)- & 8331 & 8286(23.03)- & 8345 & 8307.53(19.45)- & 7970 \\
		1100-2 & \textbf{8718} & 8649.27(18.72) & 8502 & 8464.53(18.83)- & 8543 & 8474.6(28.35)- & 8541 & 8496.87(19.98)- & 7809 \\
		1100-3 & \textbf{8830} & 8778.8(18.52) & 8674 & 8636.63(20.98)- & 8671 & 8631.87(20.41)- & 8709 & 8658.53(18.86)- & 8057 \\
		1100-4 & \textbf{8576} & 8509.63(16.37) & 8424 & 8382.83(18.39)- & 8446 & 8380.1(21.41)- & 8460 & 8406.17(20.77)- & 7983 \\
		1200-1 & \textbf{9271} & 9179.4(26.41) & 9054 & 8975.2(29.84)- & 9031 & 8975.1(30.17)- & 9044 & 9008.77(22.43)- & 8468 \\
		1200-2 & \textbf{9039} & 8958.4(28.03) & 8823 & 8745.73(31.82)- & 8794 & 8740.6(27.28)- & 8835 & 8783.57(28.94)- & 8195 \\
		1200-3 & \textbf{9302} & 9247.5(19.11) & 9060 & 9014.97(22.36)- & 9125 & 9029.67(36.57)- & 9143 & 9054.77(26.07)- & 8403 \\
		1200-4 & \textbf{9234} & 9170.83(16.59) & 9017 & 8966.17(26.24)- & 9021 & 8959.9(29.98)- & 9046 & 9001.67(20.74)- & 8177 \\
		1300-1 & \textbf{9566} & 9491.67(24.62) & 9325 & 9253.53(26.62)- & 9336 & 9261.87(34.15)- & 9404 & 9307.53(32.39)- & 8439 \\
		1300-2 & \textbf{9421} & 9358.4(17.76) & 9202 & 9144.87(25.91)- & 9190 & 9141.27(26.52)- & 9220 & 9181.03(20.5)- & 8595 \\
		1300-3 & \textbf{9376} & 9294.5(17.41) & 9146 & 9070.37(27.25)- & 9136 & 9072.97(32.7)- & 9146 & 9098.17(19.74)- & 8543 \\
		1300-4 & \textbf{9389} & 9309.63(27.44) & 9150 & 9105.93(22.93)- & 9189 & 9109.33(30.38)- & 9201 & 9142.07(26.67)- & 8384 \\
		1400-1 & \textbf{9634} & 9533.1(33.86) & 9328 & 9260(38.23)- & 9373 & 9253.37(42.43)- & 9358 & 9301.63(36.76)- & 8456 \\
		1400-2 & \textbf{10105} & 10034.5(28.55) & 9867 & 9755.37(34.76)- & 9838 & 9759.57(44.35)- & 9886 & 9794.33(30.85)- & 8738 \\
		1400-3 & \textbf{9793} & 9699.1(25.56) & 9552 & 9453.5(35.55)- & 9536 & 9454.33(36.9)- & 9579 & 9501(29.87)- & 8522 \\
		1400-4 & \textbf{9764} & 9589.5(26.46) & 9422 & 9342.33(34.25)- & 9401 & 9335.23(31.72)- & 9450 & 9381.57(30.17)- & 8458\\
		\bottomrule
	\end{tabular}
\end{table}

The experimental results for medium and large-scale instances are shown in Table 3. In the vast majority of instances, our proposed RL-GA algorithm achieves the largest detection profit. The gap between the proposed algorithm and other algorithms is obvious. According to the mean and standard deviation, it can be seen that the RL-GA algorithm has a good average performance in multiple runs, and the algorithm has good stability. It is worth mentioning that the HA algorithm with a simple structure can also search for a high-quality detection task execution plan.

The experiments of ultra-large-scale instances can more effectively reflect the solving ability of the algorithm in the case of complex constraints and large solution space. As shown in Table 4, the ultra-large-scale scheduling problem effectively utilizes the advantages of reinforcement learning in the RL-GA. It can change the strategy selection to allow the algorithm to try new search spaces continuously. Meanwhile, the elite individual retention strategy improves the local search ability of the algorithm. The performance of the other four algorithms is relatively close on the whole, and ABC performs better than other state-of-the-art algorithms in most cases.
\begin{center}
	\begin{figure}[htbp]
		\centering
		\subfigure[Performance of instance No. 300-4 ]{
			\includegraphics[width=7cm]{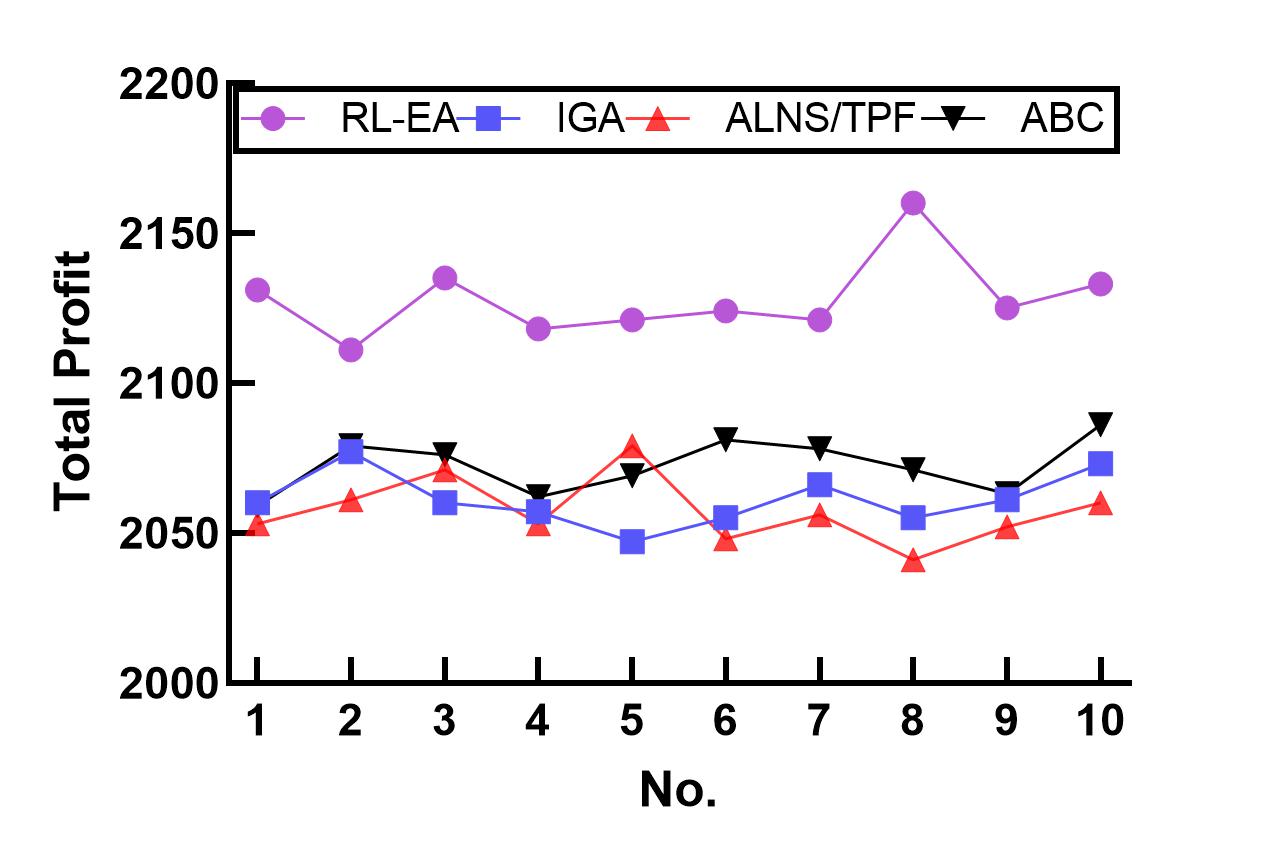}
			\label{10runsa}
		}
		\quad
		\subfigure[Performance of instance No. 600-4 ]{
			\includegraphics[width=7cm]{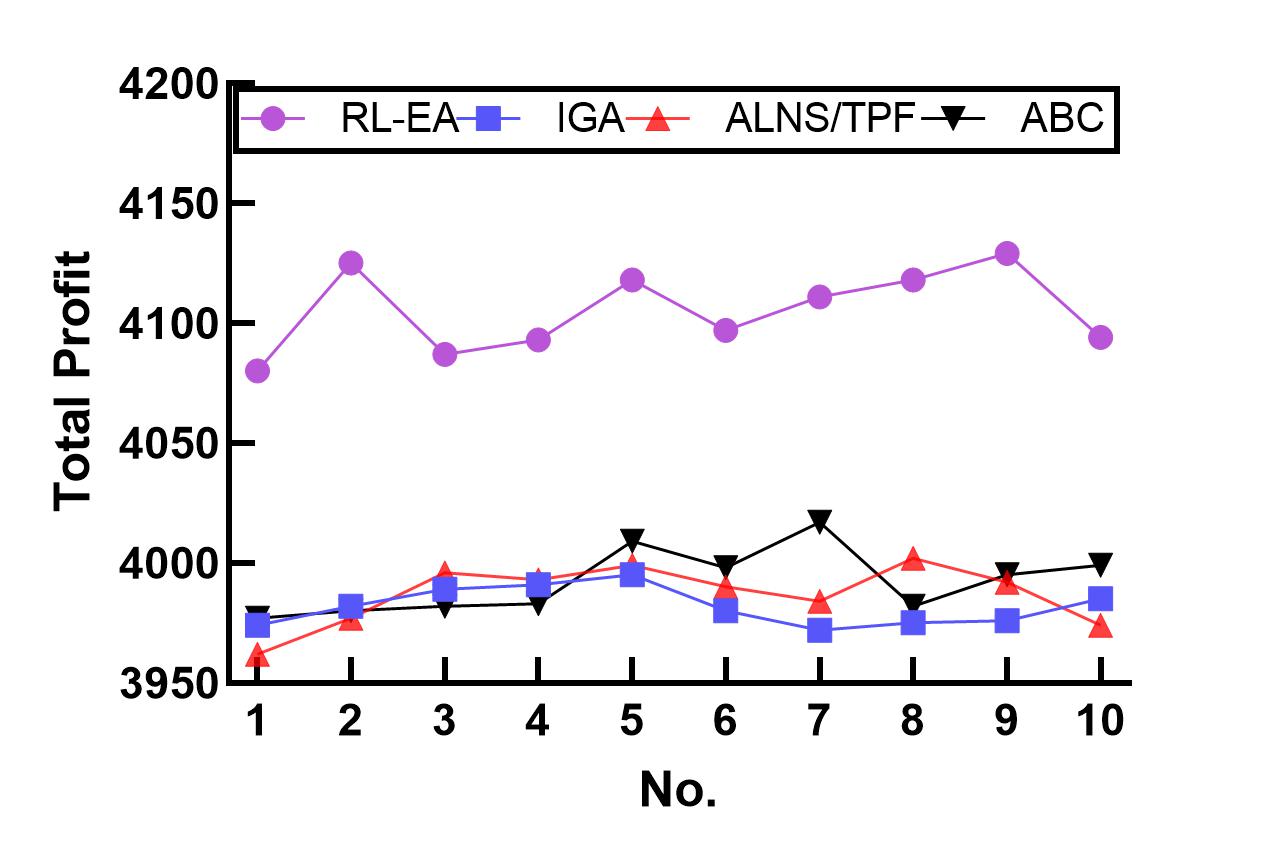}
			\label{10runsb}
		}
		\quad
		\subfigure[Performance of instance No. 800-4 ]{
			\includegraphics[width=7cm]{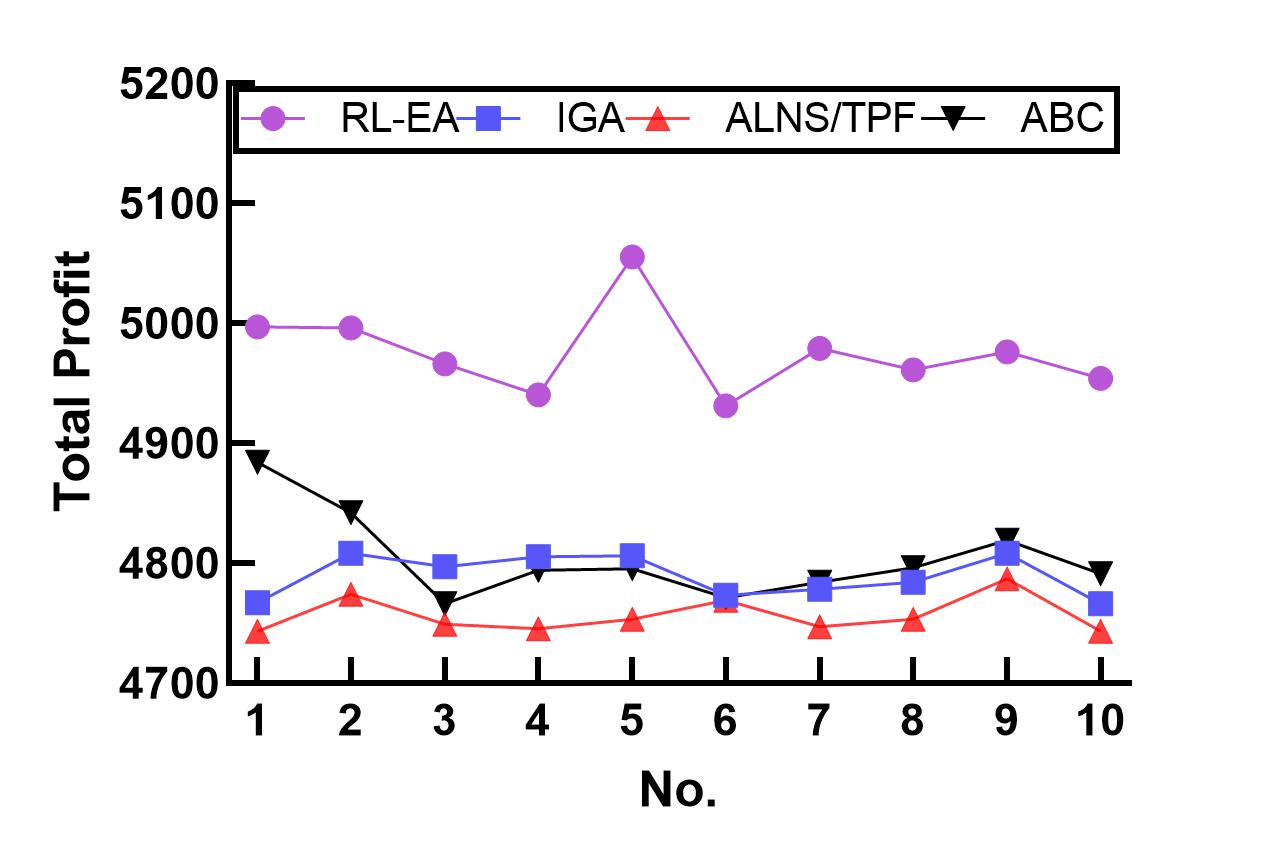}
			\label{10runsc}
		}
		\quad
		\subfigure[Performance of instance No. 1400-4 ]{
			\includegraphics[width=7cm]{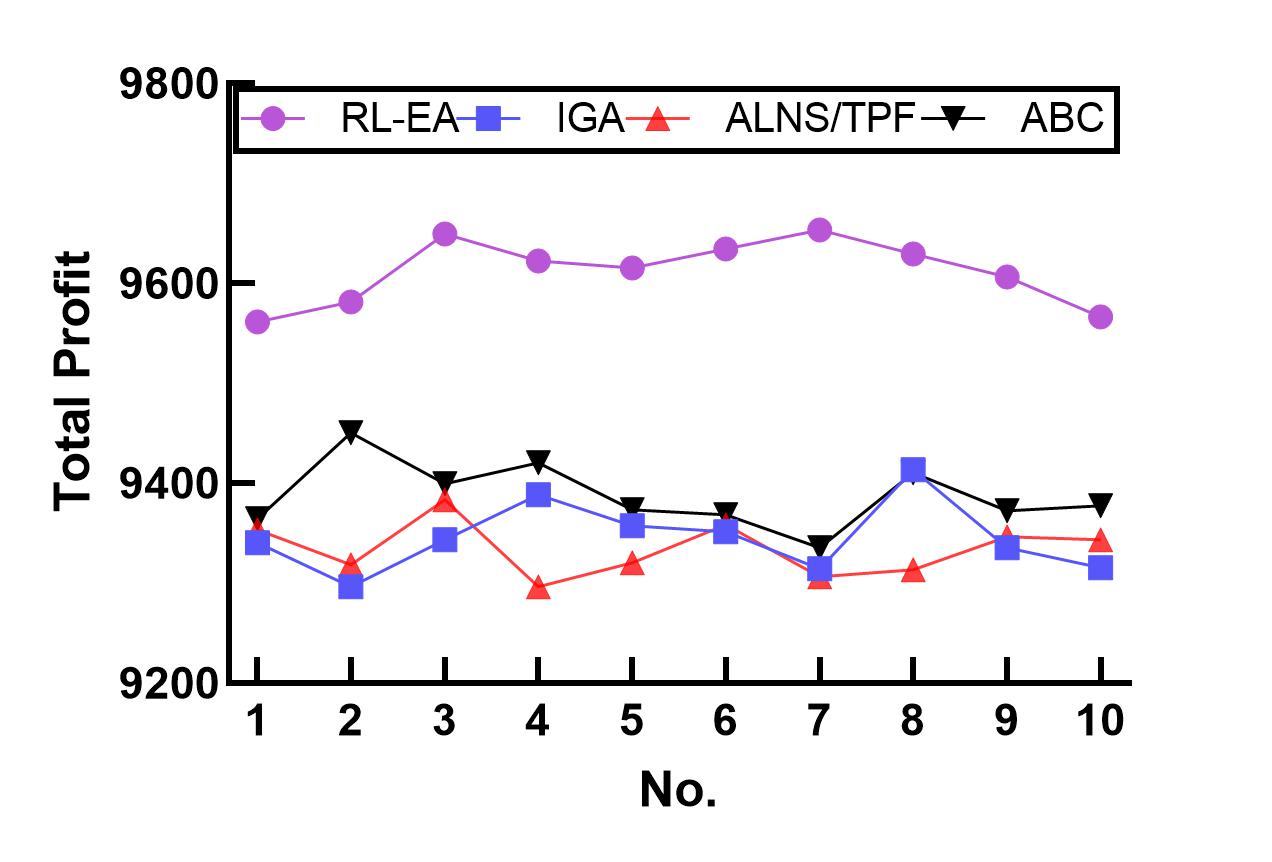}
			\label{10runsd}
		}
		\caption{Results of 10 Runs of Instances}
	\end{figure}
\end{center}

\textcolor[rgb]{0,0,0}{In addition, we also use the traditional genetic algorithm and an improved genetic algorithm called the GA\_ELUMS algorithm to test the problem-solving ability for ultra-large-scale instances, and the results are shown in Table 5.}

\begin{table}[htbp]
	\tiny
	\caption{Comparison results with traditional GA and GA\_ELUMS}
	\centering
	\begin{tabular}{ccccccc}
		\toprule
		\multirow{2}{*}{Instance} & \multicolumn{2}{l}{RL-GA} & \multicolumn{2}{l}{Traditional GA} & \multicolumn{2}{l}{GA\_ELUMS} \\
		\cmidrule(r){2-3}  \cmidrule(r){4-5} \cmidrule(r){6-7} 
		& Best & Mean(Std. Dev.) & Best & Mean(Std. Dev.) & Best & Mean(Std. Dev.) \\
		\midrule
		1000-1 & \textbf{8003}  & \textbf{7939.33(15.24)} & 7852 & 7784.07(22.44)- & 7884 & 7838.47(17.97)- \\
		1000-2 & \textbf{7821}  & \textbf{7788.6(16.31)}  & 7732 & 7677.9(21.43)-  & 7740 & 7708.6(17.54)-  \\
		1000-3 & \textbf{7907}  & \textbf{7882.27(13.95)} & 7831 & 7801.53(17.25)- & 7857 & 7831.97(19.77)- \\
		1000-4 & \textbf{8091}  & \textbf{8045.57(14.01)} & 7953 & 7912.1(20.35)-  & 7987 & 7953.37(16.34)- \\
		1100-1 & \textbf{8452}  & \textbf{8401.7(17.73)}  & 8357 & 8271.37(26.34)- & 8350 & 8307(17.78)-    \\
		1100-2 & \textbf{8718}  & \textbf{8649.27(18.72)} & 8494 & 8442.57(24.25)- & 8568 & 8500.63(23.07)- \\
		1100-3 & \textbf{8830}  & \textbf{8778.8(18.52)}  & 8660 & 8615.1(19.81)-  & 8707 & 8667(19.2)-     \\
		1100-4 & \textbf{8576}  & \textbf{8509.63(16.37)} & 8412 & 8361.33(24.02)- & 8445 & 8405.77(17.33)- \\
		1200-1 & \textbf{9271}  & \textbf{9179.4(26.41)}  & 9048 & 8961.9(33.16)-  & 9080 & 9020.13(25.91)- \\
		1200-2 & \textbf{9039}  & \textbf{8958.4(28.03)}  & 8822 & 8732.57(34.24)- & 8846 & 8790.63(30.02)- \\
		1200-3 & \textbf{9302}  & \textbf{9247.5(19.11)}  & 9064 & 9003.97(28.18)- & 9136 & 9067.43(28.53)- \\
		1200-4 & \textbf{9234}  & \textbf{9170.83(16.59)} & 9027 & 8950.7(32.27)-  & 9073 & 9015.8(23.49)-  \\
		1300-1 & \textbf{9566}  & \textbf{9491.67(24.62)} & 9306 & 9229.23(29.77)- & 9368 & 9304.37(26.95)- \\
		1300-2 & \textbf{9421}  & \textbf{9358.4(17.76)}  & 9196 & 9132.87(28.75)- & 9272 & 9201.13(26.26)- \\
		1300-3 & \textbf{9376}  & \textbf{9294.5(17.41)}  & 9143 & 9030.8(37.1)-   & 9170 & 9110.6(26.66)-  \\
		1300-4 & \textbf{9389}  & \textbf{9309.63(27.44)} & 9167 & 9090.6(33.94)-  & 9210 & 9147.4(22.58)-  \\
		1400-1 & \textbf{9634}  & \textbf{9533.1(33.86)}  & 9291 & 9235.37(34.16)- & 9381 & 9312.7(34.48)-  \\
		1400-2 & \textbf{10105} & \textbf{10034.5(28.55)} & 9790 & 9724.87(35.96)- & 9875 & 9815.73(29.2)-  \\
		1400-3 & \textbf{9793}  & \textbf{9699.1(25.56)}  & 9512 & 9436.17(41.77)- & 9591 & 9513.97(32.25)- \\
		1400-4 & \textbf{9764}  & \textbf{9589.5(26.46)}  & 9476 & 9324.37(45.95)- & 9427 & 9391.4(27.52)- \\
		\bottomrule
	\end{tabular}
\end{table}

\textcolor[rgb]{0,0,0}{From Table 5, it can be seen that the RL-GA algorithm has better planning performance compared with the traditional GA algorithm and the GA\_ELUMS algorithm \cite{b41}. The GA\_ELUMS algorithm uses a series of improved strategies containing initialization, crossover, variation, and individual selection, but these strategies are not strong enough to learn and it is difficult to adjust the search strategy by the information obtained from the search. In contrast, the Q-learning method in the RL-GA algorithm dynamically decides on the operations throughout the population evolution process. This learning method can effectively improve the algorithm's ability to search the problem solution space.}

More intuitively, the algorithm results of 300-4, 600-4, 800-4, and 1400-4 running 10 times are shown. The results are shown in Figure \ref{10runsa}, Figure \ref{10runsb}, Figure \ref{10runsc} and Figure \ref{10runsd}. As can be seen from the figure, the RL-GA algorithm can maintain good stability while obtaining high profit.

\subsubsection{Convergence Analysis}

\begin{center}
	\begin{figure}[htbp]
		\centering
		\subfigure[Convergence curves under 300 task scale]{
			\includegraphics[width=7cm]{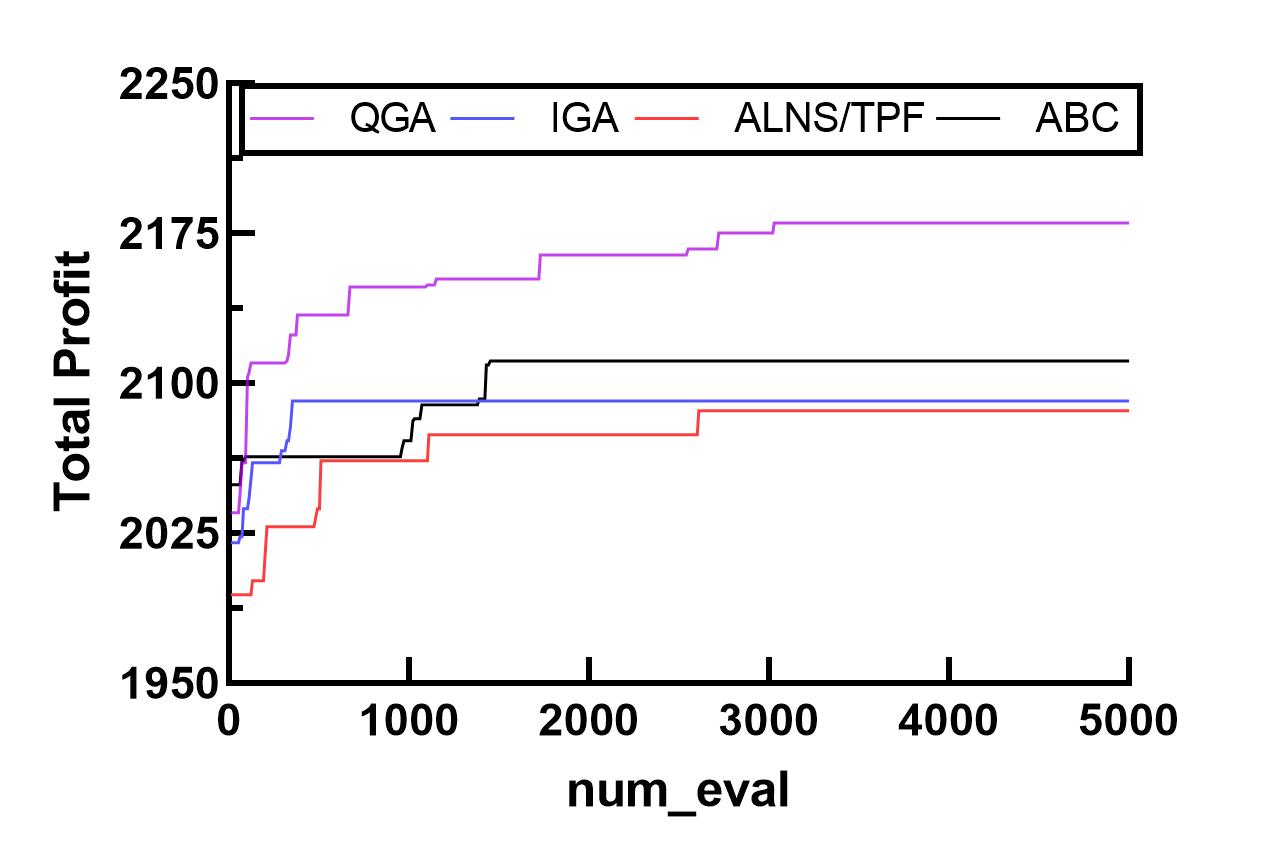}
			\label{cov1}
		}
		\quad
		\subfigure[Convergence curves under 600 task scale]{
			\includegraphics[width=7cm]{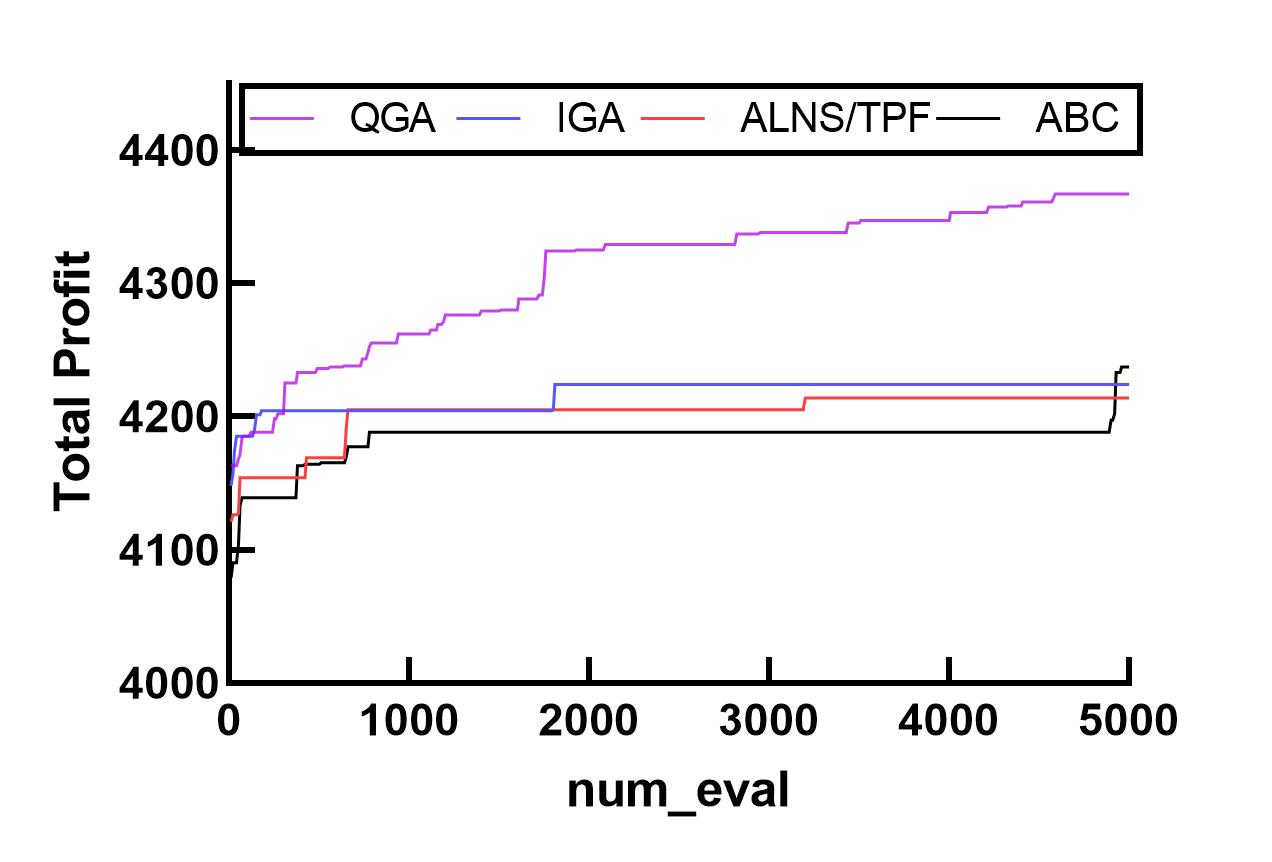}
			\label{cov2}
		}
		\quad
		\subfigure[Convergence curves under 800 task scale]{
			\includegraphics[width=7cm]{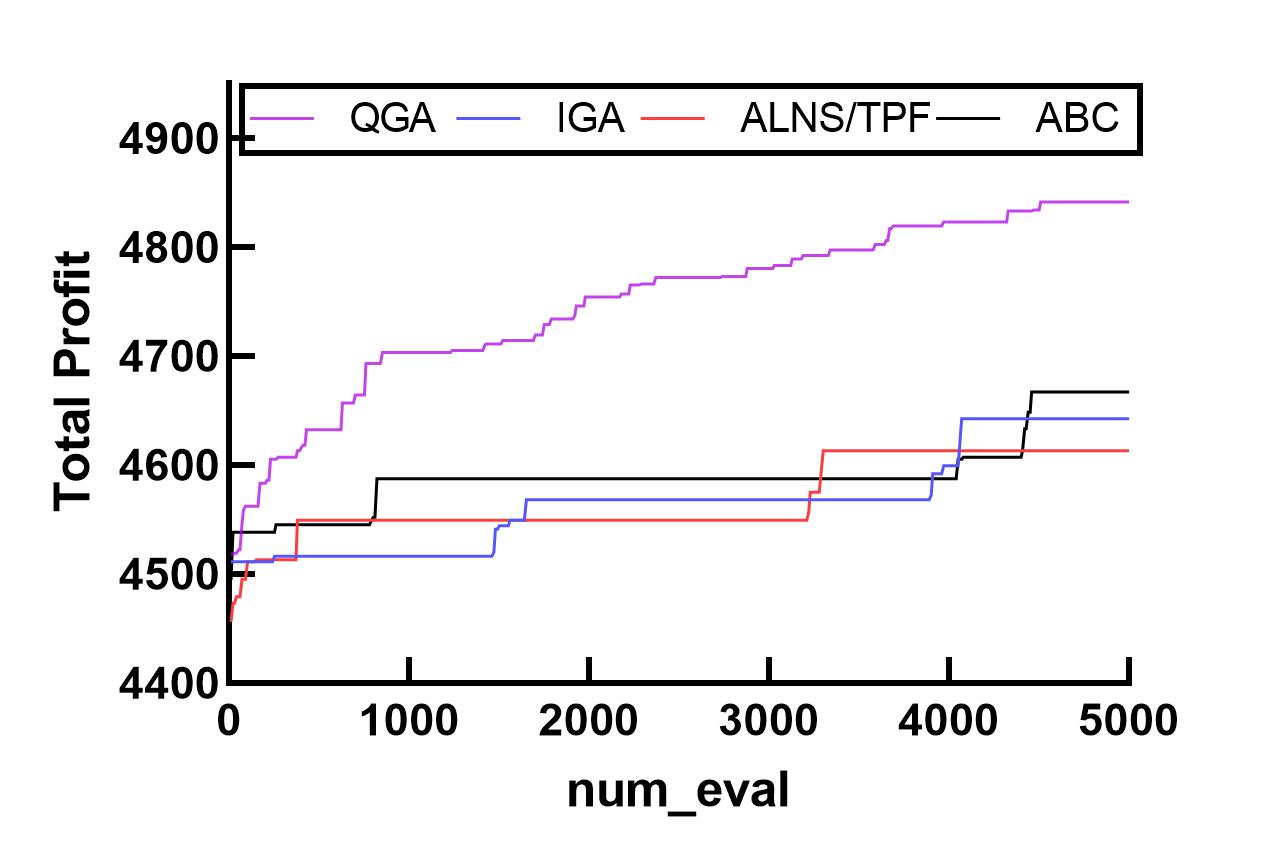}
			\label{cov3}
		}
		\quad
		\subfigure[Convergence curves under 1400 task scale]{
			\includegraphics[width=7cm]{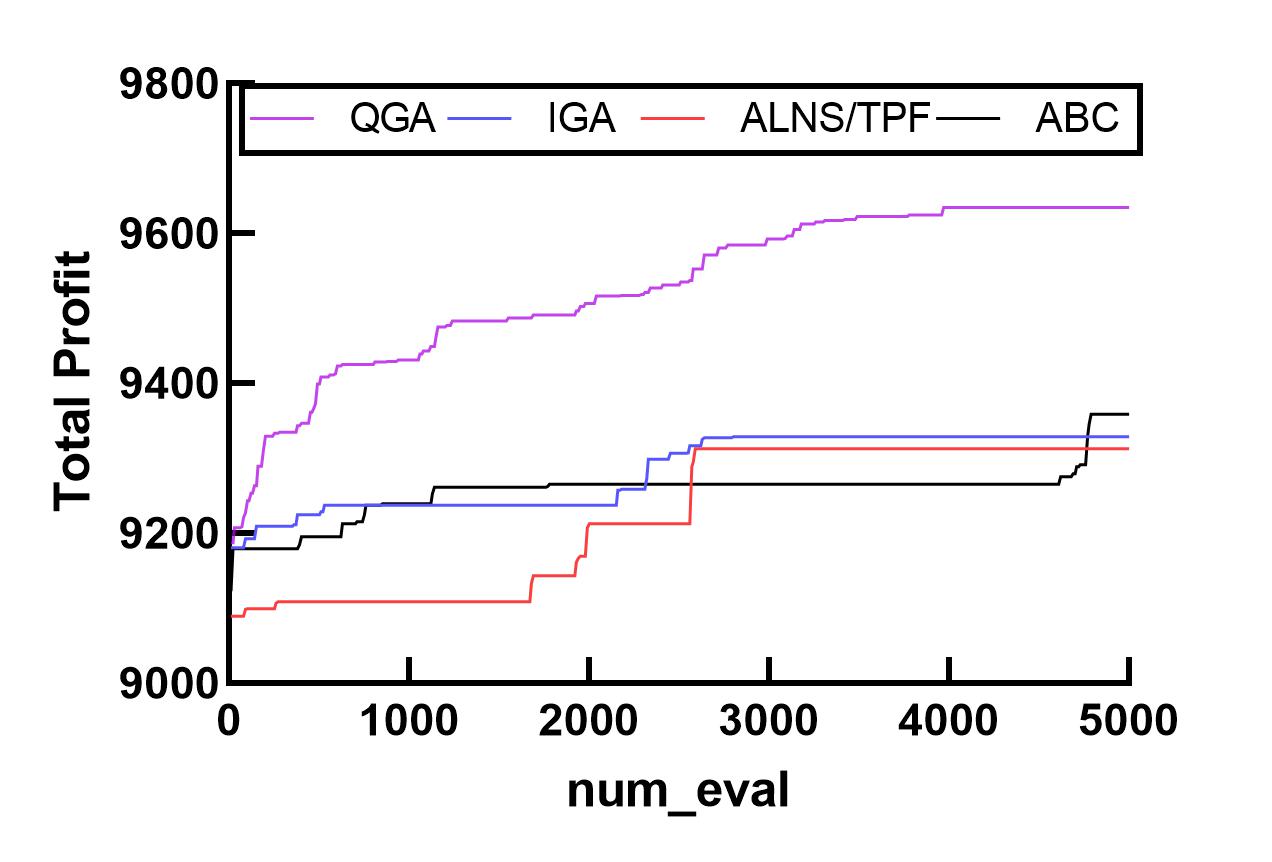}
			\label{cov4}
		}
		\caption{Convergence Curves of Different Scale Instances}
	\end{figure}
\end{center}

Figures \ref{cov1}, \ref{cov2}, \ref{cov3} and \ref{cov4} show the convergence speed of different algorithms at 300, 600, 800, and 1400 task scales, respectively. In a small-scale instance, when the number of fitness evaluations reaches 3000, the performance of the algorithm is no longer improved. When the task scale is larger, more search algebras are required for the algorithm to converge. Compared with the compared state-of-the-art algorithms, the RL-GA algorithm has a faster algorithm convergence speed and can exploit continuously.

\subsubsection{CPU Time Analysis}


We make statistics on algorithms' CPU time. The results in Table 6 are the meantime results. It can be seen from the results that the increase in time mainly comes from the increase in problem scale, and the time used for task time window selection increases significantly. When the task scale is small, the running time of the proposed algorithm is not dominant due to the existence of a series of operations of Q value calculation and action selection. When the task scale increases, the task sequence generated by the RL-GA algorithm in the front position has a high possibility of being successfully scheduled. When the time window resource cannot be scheduled, the task time window selection algorithm will not be run to try to schedule. So the computational resources consumed by the reinforcement learning in the algorithm can be ignored. It is not difficult to find that the ABC algorithm takes a lot of time, which is closely related to the complex division of labor within the population and the interaction of bees' search information. Since the environment of practical problems is often very complex, and the task scale is large, the proposed algorithm is competent for task scheduling.
\begin{table}[htbp]
	\small
	\label{Mean CPU time for Ultra-large-scale instances}
	\caption{Mean CPU Time(s) for Ultra-large-scale instances}
	\centering
	\begin{tabular}{ccccc}
		\toprule
		\textbf{Instance} & RL-GA & IGA & ALNS/TPF & ABC \\
		\midrule
		1000-1 & 2.54 & \textbf{2.53} & 2.55 & 6.54 \\
		1000-2 & \textbf{2.67} & 2.73 & 2.72 & 6.50 \\
		1000-3 & 2.61 & \textbf{2.57} & 2.58 & 6.19 \\
		1000-4 & \textbf{2.55} & 2.68 & 2.67 & 6.27 \\
		1100-1 & \textbf{2.92} & 3.02 & 3.01 & 7.27 \\
		1100-2 & \textbf{2.89} & 2.99 & 2.98 & 7.14 \\
		1100-3 & \textbf{2.91} & 2.92 & 2.94 & 6.90 \\
		1100-4 & \textbf{2.90} & \textbf{2.90} & 2.92 & 7.03 \\
		1200-1 & \textbf{3.21} & 3.28 & 3.32 & 8.08 \\
		1200-2 & \textbf{3.34} & 3.35 & 3.46 & 8.27 \\
		1200-3 & \textbf{3.40} & 3.64 & 3.52 & 8.94 \\
		1200-4 & \textbf{3.28} & 3.35 & 3.32 & 8.68 \\
		1300-1 & \textbf{3.79} & 4.03 & 3.96 & 9.47 \\
		1300-2 & \textbf{3.66} & 3.77 & 3.75 & 9.14 \\
		1300-3 & \textbf{3.54} & 3.65 & 3.61 & 9.38 \\
		1300-4 & \textbf{3.46} & 3.59 & 3.56 & 8.82 \\
		1400-1 & \textbf{4.12} & 4.19 & 4.18 & 10.82 \\
		1400-2 & \textbf{4.00} & 4.10 & 4.05 & 10.84 \\
		1400-3 & \textbf{4.21} & 4.26 & 4.40 & 10.78 \\
		1400-4 & \textbf{4.10} & 4.34 & 4.41 & 11.54\\
		\bottomrule
	\end{tabular}
\end{table}

\textcolor[rgb]{0,0,0}{It can be seen from the above experiments that RL-GA achieves better results than other comparison algorithms in terms of profit, stability, convergence speed, and CPU time. From the perspective of problem scale, the RL-GA algorithm has a good performance on large-scale problems, but its time advantage is not obvious in small-scale problems due to its Q-value evaluation and selection mechanism. From the statistical test results of the above experiments, there is a significant difference between the RL-GA algorithm proposed and other comparison algorithms at the level of $p=0.05$ in the majority of instances.}

\subsubsection{Strategy Comparison Analysis}

\begin{center}
	\begin{figure}[htbp]
		\centering
		\subfigure[Strategy comparison Results under 300 task scale]{
			\includegraphics[width=7cm]{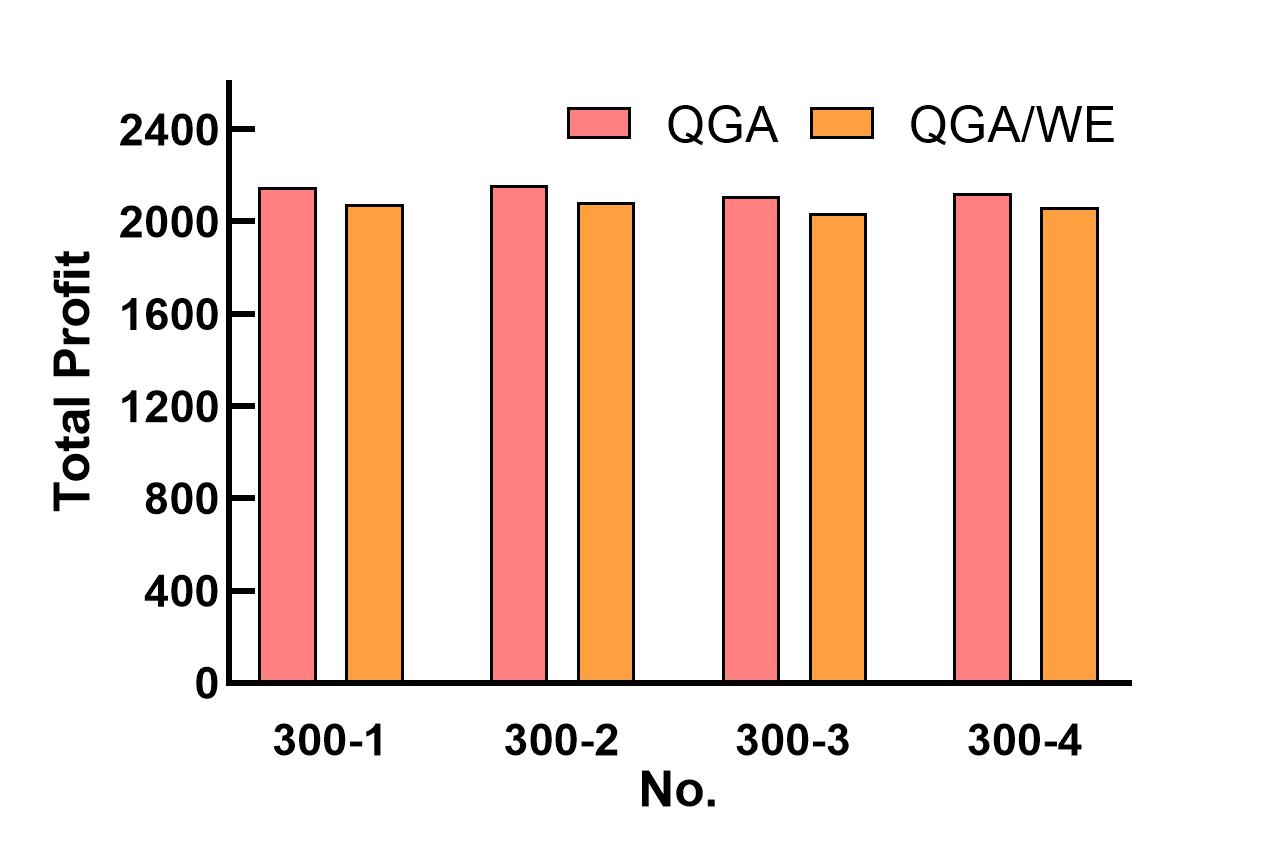}
			\label{elite1}
		}
		\quad
		\subfigure[Strategy comparison Results under 600 task scale]{
			\includegraphics[width=7cm]{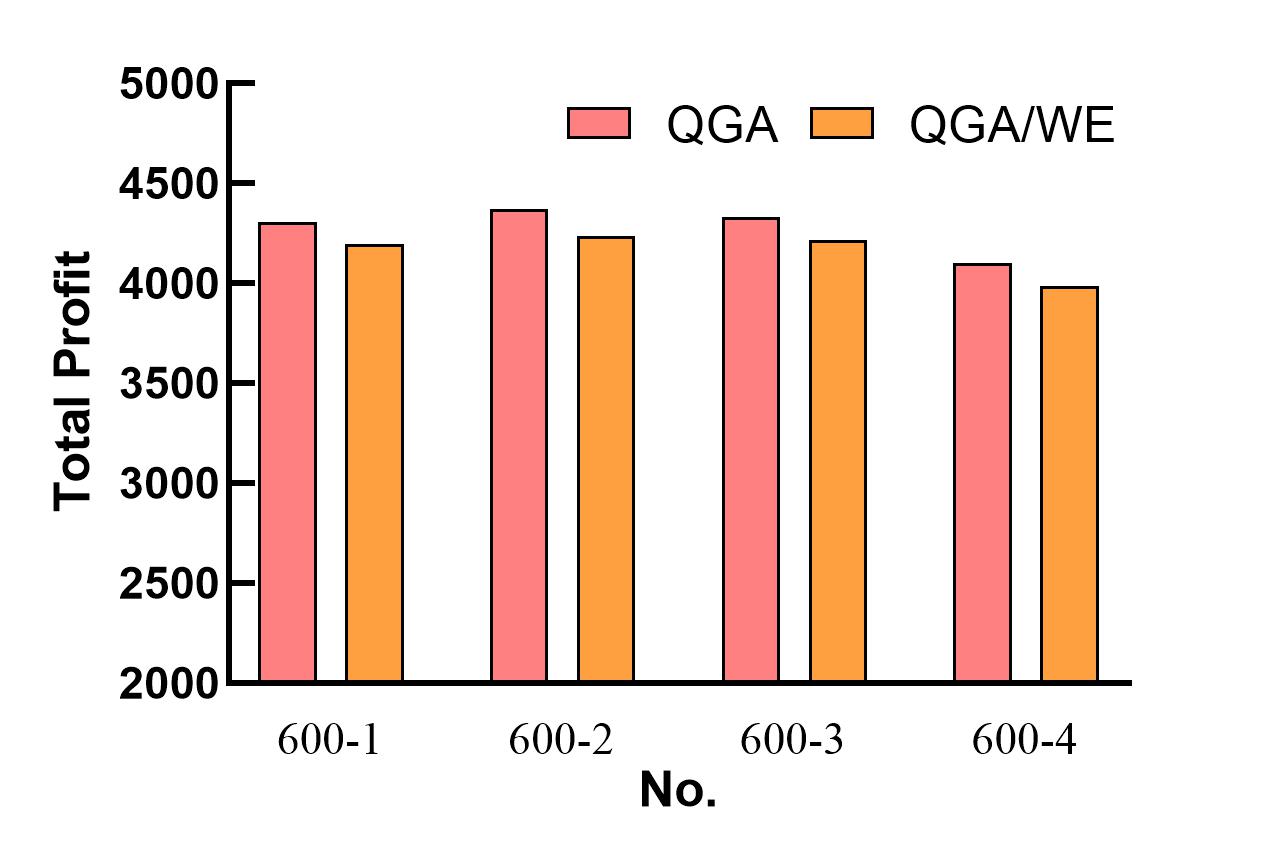}
			\label{elite2}
		}
		\quad
		\subfigure[Strategy comparison Results under 800 task scale]{
			\includegraphics[width=7cm]{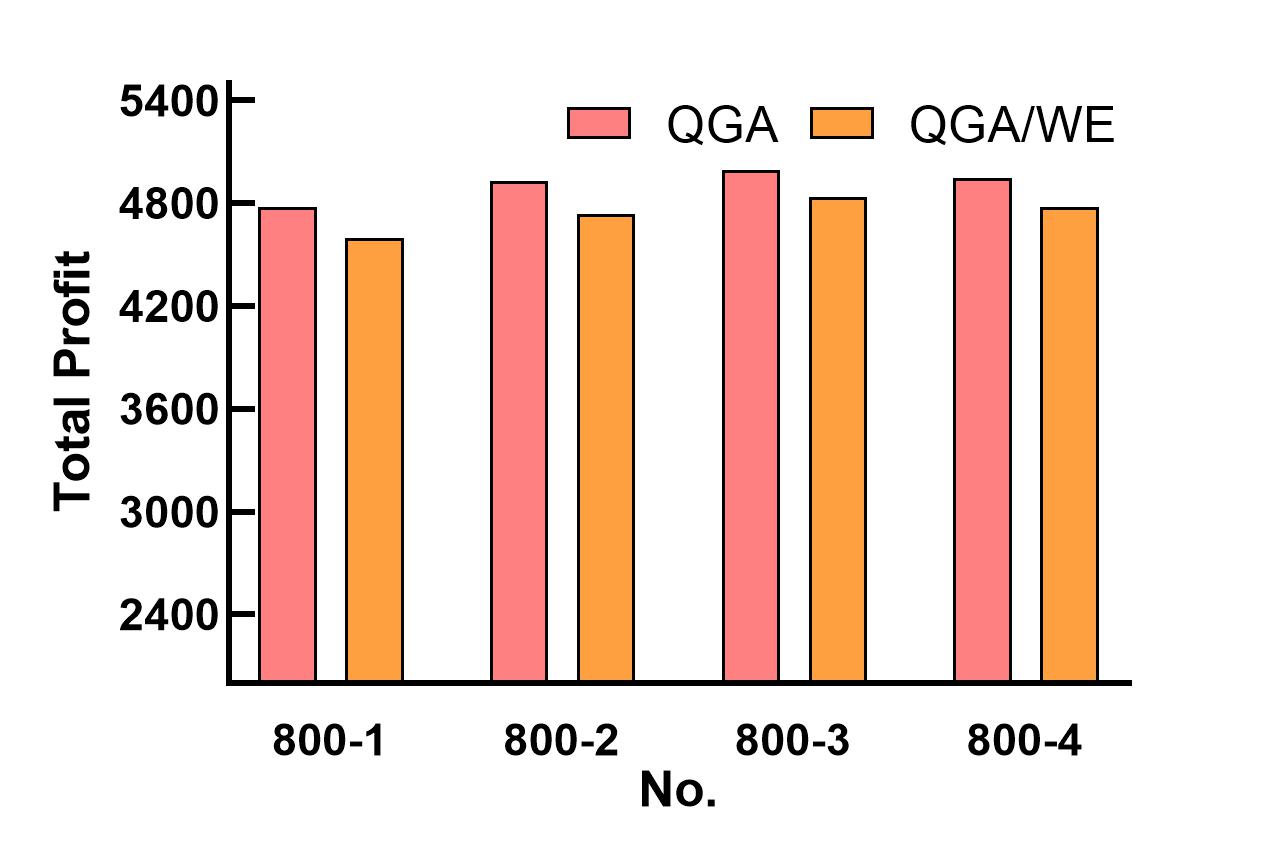}
			\label{elite3}
		}
		\quad
		\subfigure[Strategy comparison Results under 1400 task scale]{
			\includegraphics[width=7cm]{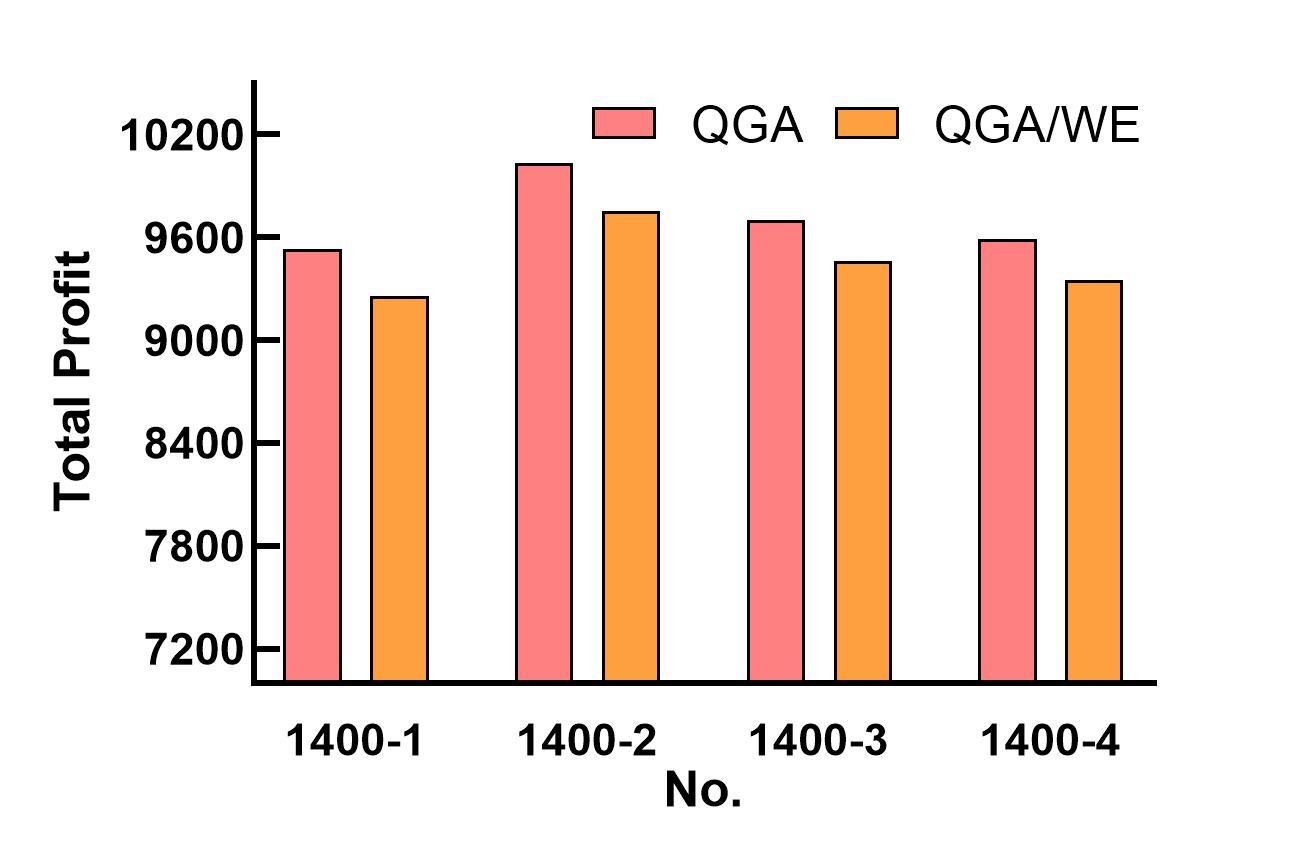}
			\label{elite4}
		}
		\caption{Strategy comparison Results of Different Scale Instances}
	\end{figure}
\end{center}

Afterward, we examine whether the elite individual retention strategy can play a role in the RL-GA algorithm. We select four instances under the task scale of 300, 600, 800, and 1400 to compare the RL-GA algorithm and the RL-GA algorithm (donated as RL-GA/WE) after removing the elite individual retention strategy. The average results of returns are shown in Figure \ref{elite1}, Figure \ref{elite2}, Figure \ref{elite3}, and Figure \ref{elite4}, and it can be seen that the elite individual retention strategy makes sense for RL-GA. The RL-GA algorithm that contains an elite individual retention strategy can always achieve higher profit.

\textcolor[rgb]{0,0,0}{In the following, we will use a case study to verify the performance of the RL-GA algorithm in a real project.}

\subsubsection{Case Study}

\textcolor[rgb]{0,0,0}{The satellite scheduling system contains a central node and several user nodes, where the user nodes are responsible for submitting applications and the central node deploys mission planning software for generating execution plans. After the user applies, the central node will pre-process it based on a series of information such as location, satellite orbit, and other requirements. The pre-processing will remove duplicate tasks, merge similar tasks, and split tasks that cannot be detected at once. Preprocessing also calculates the visible time window of each task. Once the preprocessing is complete, the input data needed for scheduling is obtained, including satellites, tasks, and time windows.}

\begin{center}
	\begin{figure}[htbp]
		\centering
		\subfigure[Detection Profit]{
			\includegraphics[width=7cm]{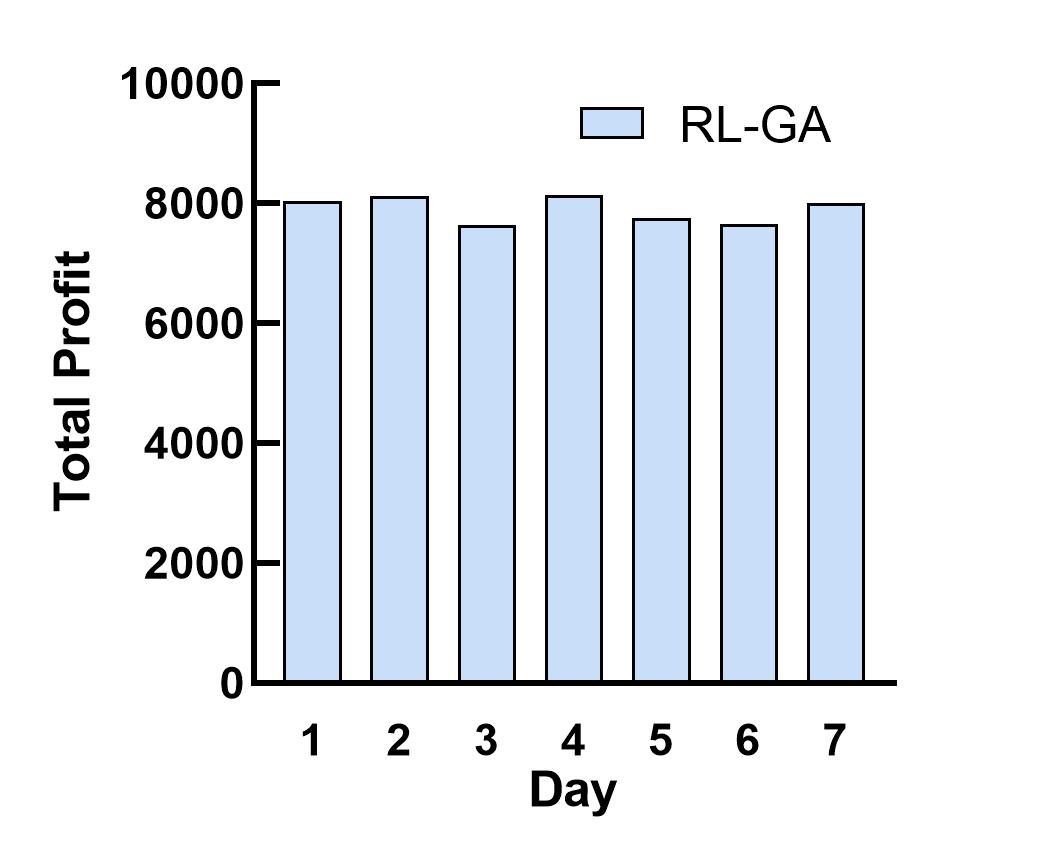}
			\label{case1}
		}
		\quad
		\subfigure[Task Completion Rate]{
			\includegraphics[width=7cm]{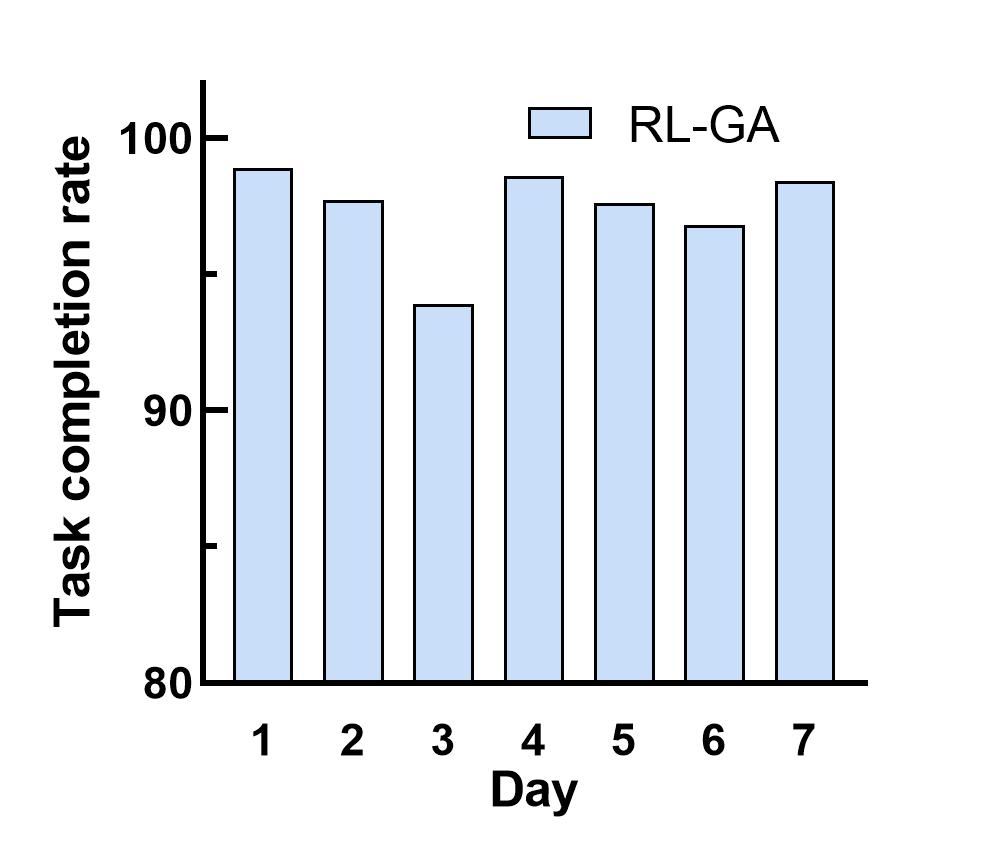}
			\label{case2}
		}
		\caption{Scheduling Results of the Satellite Scheduling System}
		\label{case}
	\end{figure}
\end{center}

\textcolor[rgb]{0,0,0}{The satellite scheduling system tends to use a week as a scheduling cycle.  We use the RL-GA algorithm proposed in this paper to plan each day's tasks separately, obtain the execution plan for a single day, and then accumulate to get the execution plan for the whole week. Here, we let the system plan 1000 tasks per day, and the scheduling results are shown in Figure \ref{case1}. The task completion rate is also a factor considered in practical applications, and the successful task execution rate for each day is given in Figure \ref{case2}.}

\begin{center}
	\begin{figure}[htp]
		\centerline{\includegraphics[scale=0.6,trim=0 0 0 0]{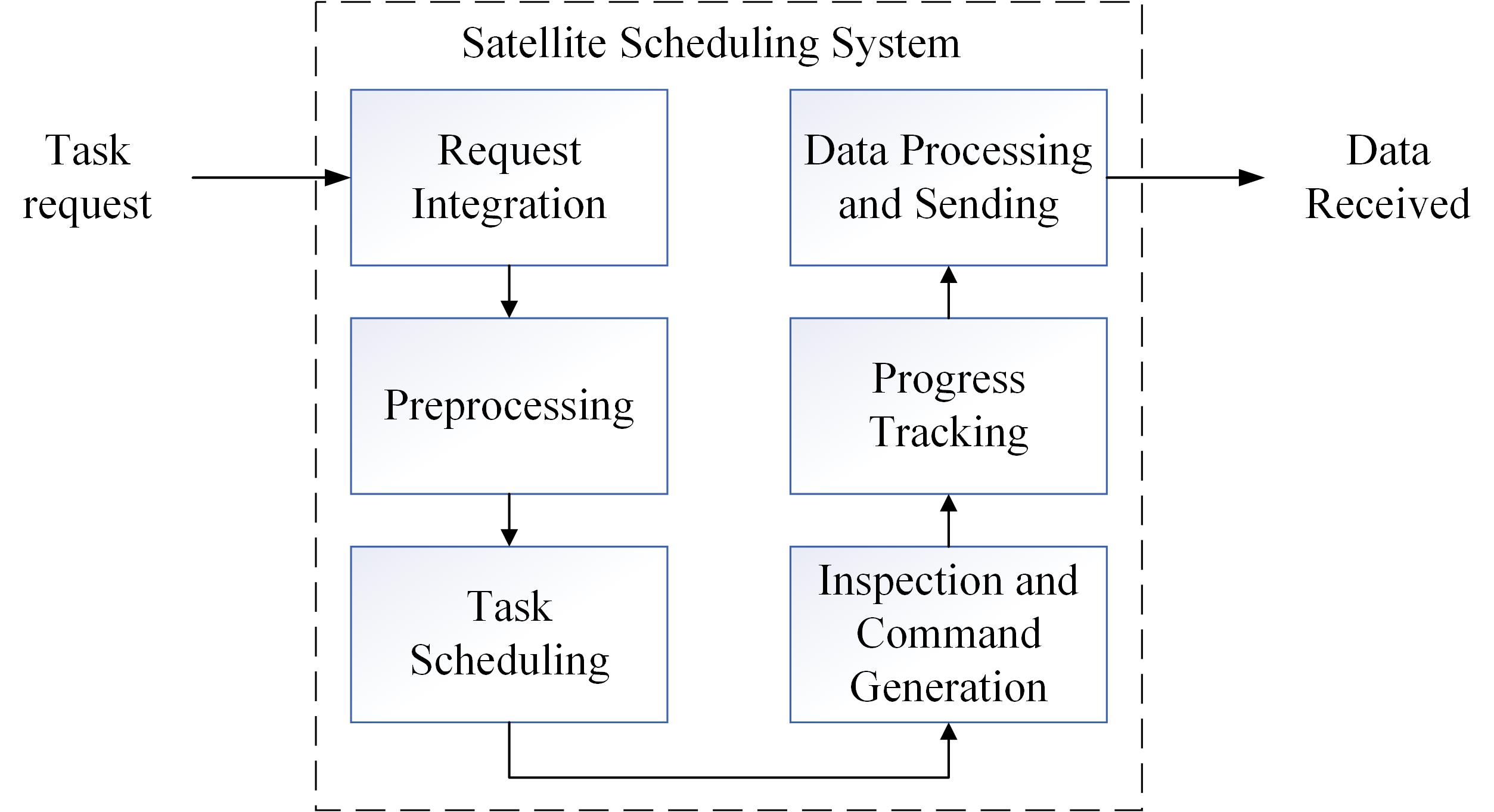}}
		\caption{Processing Flow of the Satellite Scheduling System}
		\label{pic processing flow of the satellite scheduling system}
	\end{figure}
\end{center}

\textcolor[rgb]{0,0,0}{As shown in Figure \ref{case}, the satellite scheduling system obtains a steady daily profit and less than 10\% of the tasks cannot be completed. After generating the plan, the satellite scheduling system will check the correctness of the generated solution and pass the test by generating satellite commands to the satellites that need to execute tasks through ground stations. The satellite will execute each task according to the start and end time set in the plan and transmit the data obtained to the ground station through data downlink. The data is processed accordingly and distributed to the user who submitted the request. The processing flow of the satellite scheduling system is shown in Figure \ref{pic processing flow of the satellite scheduling system}.}

\textcolor[rgb]{0,0,0}{From the above results, we can see that our proposed algorithm can solve the EDSSP problem well and verify the effectiveness of the algorithm from several aspects. the RL-GA algorithm not only can get high-quality solutions but also can solve the problem quickly. Large-scale instances and ultra-large-scale instances further prove the ability of the RL-GA algorithm to solve practical problems. The results of real cases verify that the algorithm can maintain good planning performance throughout the system's multi-day operation and is available for long-term use.}

\section{Conclusion}

\textcolor[rgb]{0,0,0}{In this paper, we propose a reinforcement learning-based genetic algorithm to solve the EDSSP problem. The genetic algorithm based on reinforcement learning uses a new combination method of Q-learning and GA algorithm, which allows the algorithm to select operators autonomously through adaptive learning, giving the algorithm a strong ability to adapt to different scenarios. The agent can choose effective actions based on the interaction with the environment during the search process to improve search performance. The population search process in the evolutionary algorithm is guided by reinforcement learning, which allows each individual to select an appropriate strategy according to the reward and Q value. We construct a $<$state, action$>$ combination method according to the search performance and update the corresponding value after each evolution is completed. The elite individual retention strategy also used in the algorithm can improve search performance. It can be seen from the experiments that our proposed algorithm has obvious advantages in solving the EDSSP problem. }

\textcolor[rgb]{0,0,0}{LA-GA effectively solves the EDSSP problem and enables the satellite scheduling system to obtain a high-quality mission execution plan in a short time. The proposed algorithm can effectively cope with the increase in the number of satellites and tasks. In addition, the problem-solving idea is also applicable to other combinatorial optimization problems with order dependence.}

In future research, we will try to reduce the computational cost of the Q-value evaluation mechanism to further accelerate the search speed without affecting the performance of the algorithm. Other learning methods will also be considered, such as transfer learning, deep learning, and others. In addition, we will consider applying learning-based approaches to the online scheduling problem. This type of scheduling problem has higher requirements for the algorithm.

\section{Declaration of Competing Interest}

The authors declare that they have no known competing financial interests or personal relationships that could have appeared to influence the work reported in this paper.

\section{Acknowledgements}
This work was supported by the National Natural Science Foundation of China (71901213, 61773120, 72001212), the Special Projects in Key Fields of Universities in Guangdong (2021ZDZX1019), and the Hunan Provincial Innovation Foundation For Postgraduate (CX20200585).

Thanks to the editor and reviewers for their valuable comments. We also thank Dr. Luona Wei and Dr. Yongming He for their guidance on the method.

Luona Wei and Yanjie Song contribute equally to this article.

\end{document}